\documentclass{article}
\usepackage{arxiv}
\usepackage[utf8]{inputenc} 
\usepackage[T1]{fontenc}    
\usepackage{hyperref}       
\usepackage{url}            
\usepackage{booktabs}       
\usepackage{amsfonts}       
\usepackage{nicefrac}       
\usepackage{microtype}      
\usepackage{lipsum}		
\usepackage{graphicx}
\usepackage[numbers]{natbib}
\usepackage{doi}
\usepackage{amssymb, mathtools, caption, subcaption, algorithm}
\usepackage[noend]{algpseudocode}

\title{Bayesian deep learning framework for uncertainty quantification in high dimensions}

\author{Jeahan Jung \\
	Department of Mathematics\\
	Pohang University of Science and Technology (POSTECH)\\
	Pohang 37673, Republic of Korea \\
	\texttt{obok13@postech.ac.kr} \\
	\And
        Minseok Choi \\
	Department of Mathematics\\
	Pohang University of Science and Technology (POSTECH)\\
	Pohang 37673, Republic of Korea \\
	\texttt{mchoi@postech.ac.kr}
}

\date{}

\renewcommand{\shorttitle}{\textit{arXiv} Template}

\hypersetup{
pdftitle={A template for the arxiv style},
pdfsubject={cs.AI},
pdfauthor={Jeahan Jung, Minseok Choi},
pdfkeywords={Bayesian neural network, Hamiltonian Monte Carlo, uncertainty quantification, high dimension},
}

\begin{document}
\maketitle

\begin{abstract}
We develop a novel deep learning method for uncertainty quantification in
stochastic partial differential equations based on Bayesian neural network
(BNN) and Hamiltonian Monte Carlo (HMC). A BNN efficiently learns the
posterior distribution of the parameters in deep neural networks by
performing Bayesian inference on the network parameters. The
posterior distribution is efficiently sampled using HMC to quantify
uncertainties in the system. Several numerical examples are shown for both
forward and inverse problems in high dimension to demonstrate the
effectiveness of the proposed method for uncertainty quantification. These also
show promising results that the computational
cost is almost independent of the dimension of the problem demonstrating the
potential of the method for tackling the so-called curse of dimensionality. 
\end{abstract}

\keywords{Bayesian neural network \and Hamiltonian Monte Carlo \and uncertainty quantification \and high dimension}

\section{Introduction}

Recent research into deep learning (DL) has demonstrated the effectiveness of neural networks (NNs) when applied to partial differential equations (PDEs). The authors in \cite{zhuang2021learned,bar2019learning} proposed a method based on numerical discretization schemes to learn optimized approximations; this approach uses NNs to estimate spatial derivatives that best satisfy the equations on a low-resolution grid. In \cite{chudomelka2020deep,rudd2015constrained}, NNs are used to predict coefficients of the Galerkin approximation of the solution. Along the advances in automatic differentiation techniques, a novel method of modeling the solution, called a physics-informed NN (PINN) has been proposed \cite{raissi2020hidden,raissi2019physics,kharazmi2019variational,raissi2018hidden}. The effectiveness of PINNs has been demonstrated in a variety of PDE problems including the forward and inverse problems. The authors in \cite{wang2021learning,li2020fourier} proposed an approach to learn a solution operator of PDEs to map initial conditions, boundary conditions, or source terms, to output solutions. The use of NNs for PDE has also been proposed two decades ago \cite{lagaris1998artificial,lagaris2000neural,dissanayake1994neural,meade1994solution}, but use of large-scale NNs was computationally prohibitive at the time. Advances in computer hardware such as GPUs and recent Python libraries such as Tensorflow \cite{tensorflow2015-whitepaper} and Pytorch \cite{NEURIPS2019_9015} provide tools for DL-based PDE solver.

Uncertainty quantification in PDE problems is necessary to assess the reliability of obtained solutions. The quantified uncertainties prevent overconfidence in the accuracy of the prediction and provide probabilistic information about the confidence interval. The uncertainties have several sources, including PDE parameters, boundary conditions, and initial conditions. PDEs with this type of uncertainty are described as stochastic PDEs (SPDEs). They arise in many physical and engineering problems of random phenomena, but they often lead to high-dimensional problems, so the computational cost to solve them can be prohibitive. Recent research on DL has proposed effective ways to overcome this problem. In \cite{tripathy2018deep,zhu2019physics,zhu2018bayesian,mo2019deepa,mo2019deepb}, the authors dealt with the SPDE of Darcy's law where the randomness is induced by heterogeneous porous media. NNs are used to learn the operator between the solution and permeability, and achieve good performance in high-dimensional problems. The authors in \cite{yang2020physics,yang2019adversarial} used a generative adversarial network to provide a framework to solve both forward and inverse problems of SPDEs with data collected from a limited number of sensors; this method could solve high-dimensional problems with low-polynomial growth in computational cost, in principle. In \cite{guo2022normalizing}, a normalizing flow is used to generate stochastic solution samples in a problem setup similar to \cite{yang2020physics}. The authors in \cite{zhang2019quantifying} proposed a method that extends the polynomial chaos \cite{xiu2002wiener}, the standard method for solving SPDE; the coefficients of polynomial chaos are modeled using NNs. Similarly, the authors in \cite{zhang2020learning} expanded the dynamically-orthogonal (DO) and bi-orthogonal (BO) methods \cite{choi2014equivalence} to solve time-dependent SPDEs by using an NN to model each DO and BO component. The authors in \cite{karumuri2020simulator,nabian2018deep} approximated solutions of SPDEs by using NNs that take as input physical variables and random variables that characterize the randomness of the SPDEs.

In this paper, we use a Bayesian NN (BNN) to quantify uncertainties in SPDE problems. A BNN is an NN model that learns a distribution of its parameters based on given data \cite{neal2012bayesian,graves2011practical,blundell2015weight,gal2016dropout,welling2011bayesian,liu2016stein}. A BNN simulates the different possibilities of the model, and therefore is a natural approach to quantifying uncertainty. In addition, BNNs outperform general NNs that provide point estimates of parameters by avoiding overfitting to noisy and sparse data \cite{yang2021b,sun2020physics,blundell2015weight}. BNNs have been used to provide a reasonable uncertainty of the PDE solution depending on the magnitude of the data noise and locations of the sensors collecting data \cite{yang2021b,sun2020physics}. In \cite{zhu2018bayesian}, BNNs are used to learn solution operators of Darcy's law by using image-to-image regression. Our proposed method uses BNN to learn SPDE solutions and aims to solve a wider range of SPDE problems. We achieve this goal by encoding the governing physical law of the SPDE into the posterior distribution of the network parameters using automatic differentiation, the performing Bayesian inference on the network parameters based on the given data, and finally using the Hamiltonian Monte Carlo (HMC) method \cite{duane1987hybrid,neal2012bayesian,betancourt2017conceptual} to generate samples of the stochastic solution. This approach provides a data-driven framework for solving both forward and inverse problems of SPDEs. The proposed method can solve high-dimensional SPDEs without increasing the computational cost. Our method is implemented in various SPDE problems including high-dimensional forward and inverse problems to illustrate the efficacy of the proposed method.

The remainder of the paper is organized as follows. In Section \ref{sec:method}, we set up the forward and inverse problems of SPDEs and describe the BNN methodology for solving them. In Section \ref{sec:example}, we show and discuss various numerical simulation results of our proposed method. In Section \ref{sec:conclusion}, we conclude and summarize.

\section{Methods}\label{sec:method}

In this section, we formulate the SPDE problems of concern and introduce the BNN method.

\subsection{Problem setup}

Let $(\Omega,\mathcal{F},\mathcal{P})$ be a probability space where $\Omega$ is the sample space, $\mathcal{F}$ is its $\sigma$-algebra, and $\mathcal{P}$ is its probability measure. We consider the stochastic boundary-value problem
\begin{equation}\label{eqn:BVP}
\begin{alignedat}{3}
\mathcal{L}[u(x,\omega),x;k(x,\omega)] &= f(x,\omega), &&\;\;\text{for $x\in D,\omega \in \Omega$}, \\
\mathcal{B}[u(x,\omega),x;k(x,\omega)] &= g(x,\omega), &&\;\;\text{for $x\in \partial D,\omega \in \Omega$},
\end{alignedat}
\end{equation}
where $\mathcal{L}$ is a differential operator, $\mathcal{B}$ is a boundary operator, $D \subset \mathbb{R}^n$ is a physical domain, $u$ is the solution of the SPDE, $\omega \in \Omega$ is a random event, and $k$ is a random physical parameter that characterizes the randomness of $\mathcal{L}$ and $\mathcal{B}$.

The goal of the forward problem is to compute the stochastic solution $u$ given parameter $k$. We assume that the values of $f,g$, and $k$ are collected on the spatial sensors $\{x^{f,i}\}_{i=1}^{N_f} \subset D,\{x^{g,i}\}_{i=1}^{N_g} \subset \partial D$, and $\{x^{k,i}\}_{i=1}^{N_k} \subset D$, respectively. Because $f,g$, and $k$ are random processes, their measurements on the sensors have different values for each random event $\omega \in \Omega$. Suppose that given dataset $\mathcal{D}$ is a limited number of measurements of $f,g$, and $k$ on the sensors for different random events, i.e., $\mathcal{D} = \{[\mathbf{f}(\omega_j),\mathbf{g}(\omega_j),\mathbf{k}(\omega_j)]\}_{j=1}^{N_\omega}$ where
\begin{align*}
    \mathbf{f}(\omega) &= [f(x^{f,1},\omega),\cdots,f(x^{f,N_f},\omega)], \\
    \mathbf{g}(\omega) &= [g(x^{g,1},\omega),\cdots,g(x^{g,N_g},\omega)], \\
    \mathbf{k}(\omega) &= [k(x^{k,1},\omega),\cdots,k(x^{k,N_k},\omega)].
\end{align*}
Measurements $[\mathbf{f}(\omega_j),\mathbf{g}(\omega_j),\mathbf{k}(\omega_j)]$ in $\mathcal{D}$ are called snapshots, which provide probabilistic information about random vectors $\mathbf{f},\mathbf{g}$, and $\mathbf{k}$.

Conversely, the purpose of the inverse problem is to infer unknown parameter $k$ given data of $u,f$, and $g$. The data of $u$ are assumed to be collected on the spatial sensors $\{x^{u,i}\}_{i=1}^{N_u} \subset D$, and the data becomes $\mathcal{D} = \{[\mathbf{f}(\omega_j),\mathbf{g}(\omega_j),\mathbf{u}(\omega_j)]\}_{j=1}^{N_\omega}$ where
\begin{equation*}
\mathbf{u}(\omega) = [u(x^{u,1},\omega),\cdots,u(x^{u,N_u},\omega)].
\end{equation*}

\subsection{Data-driven modeling using Bayesian NN}

We first describe the data-driven method (Algorithm \ref{alg:BNN_method}) for the forward problem. Let $K(x,\theta)$ and $U(x,\theta)$ denote outputs of an NN model for the parameter $k$ and solution $u$, respectively, where $\theta$ denotes the parameters of the model. The distribution of $\theta$ is determined using Bayesian inference, which updates the prior distribution to the posterior distribution as additional data or information become available. To be precise, Bayes' rule states that the posterior distribution given data is the renormalized pointwise product of the prior and likelihood distributions.

The prior distribution $p(\theta)$ is elicited from previous experiments or the subjective belief of the user. Likelihood distribution $p(\mathcal{D}|\theta)$ assigns probabilistic prediction to the observations for each parameter $\theta$. Let $F$ and $G$ be the representations of $f$ and $g$, respectively, obtained by substituting the BNN outputs $K$ and $U$ in operators $\mathcal{L}$ and $\mathcal{B}$, i.e.,
\begin{align*}
F(x,\theta) &= \mathcal{L}[U(x,\theta),x;K(x,\theta)], \\
G(x,\theta) &= \mathcal{B}[U(x,\theta),x;K(x,\theta)].
\end{align*}
We can exploit automatic differentiation to apply differential operator $\mathcal{L}$ to $K$ and $U$. The likelihood distribution is then
\begin{equation*}
p(\mathcal{D}|\theta) = p_{\mathbf{f},\mathbf{g},\mathbf{k}}([\mathbf{F}(\theta),\mathbf{G}(\theta),\mathbf{K}(\theta)]),
\end{equation*}
where
\begin{align*}
\mathbf{F}(\theta) &= [F(x^{f,1},\theta),\cdots,F(x^{f,N_f},\theta)], \\
\mathbf{G}(\theta) &= [G(x^{g,1},\theta),\cdots,G(x^{g,N_g},\theta)], \\
\mathbf{K}(\theta) &= [K(x^{k,1},\theta),\cdots,K(x^{k,N_k},\theta)],
\end{align*}
and $p_{\mathbf{f},\mathbf{g},\mathbf{k}}$ is a joint probability density function (PDF) of $\mathbf{f},\mathbf{g}$, and $\mathbf{k}$. In our data-driven setting, the exact form of $p_{\mathbf{f},\mathbf{g},\mathbf{k}}$ is not given, but we can estimate it from the snapshots by using density estimation methods including kernel density estimation \cite{parzen1962estimation,silverman2018density}, normalizing flow \cite{rezende2015variational,dinh2016density,papamakarios2017masked,kingma2016improved}, and Gaussian mixture model \cite{paalanen2006feature,bilmes1998gentle,yu2011solving}. In this paper, we use a Gaussian mixture model to estimate $p_{\mathbf{f},\mathbf{g},\mathbf{k}}$, i.e.,
\begin{equation}\label{eqn:gmm}
p_{\mathbf{f},\mathbf{g},\mathbf{k}} = \sum_{i=1}^{N_c} w_i \mathcal{N}(\mathbf{m}_i,\mathbf{\Sigma}_i),
\end{equation}
where $w_i,\mathbf{m}_i$, and $\mathbf{\Sigma}_i$ are respectively the weight, mean, and covariance of the $i$-th component of the mixture model for $i=1,\cdots,N_c$ computed by the expectation-maximization algorithm \cite{bilmes1998gentle,yu2011solving}. Finally, the posterior distribution of $\theta$ given dataset $\mathcal{D}$ is derived by Bayes' rule
\begin{equation}\label{eqn:Bayes_rule}
p(\theta |\mathcal{D})=\frac{p(\mathcal{D}|\theta)p(\theta)}{\int p(\mathcal{D}|\theta)p(\theta)d\theta} \propto p(\mathcal{D}|\theta)p(\theta).
\end{equation}

Given the posterior distribution, the mean and variance of the solution for new input $x^*$ are
\begin{equation}\label{eqn:mean_and_variance}
\begin{alignedat}{2}
\mathbb{E}U(x^*) &= \int U(x^*,\theta)p(\theta |\mathcal{D})d\theta, \\
\text{Var}U(x^*) &= \int (U(x^*,\theta)-\mathbb{E}U^*)^2p(\theta |\mathcal{D})d\theta.
\end{alignedat}
\end{equation}
The integrals in (\ref{eqn:mean_and_variance}) can be estimated using a Monte Carlo estimator
\begin{equation}\label{eqn:mean_and_variance_estimate}
\begin{alignedat}{2}
\mathbb{E}U(x^*) &\approx \frac{1}{N} \sum_{i=1}^N U(x^*,\theta^i), \\
\text{Var}U(x^*) &\approx \frac{1}{N} \sum_{i=1}^N (U(x^*,\theta^i) - \mathbb{E}U^*)^2,
\end{alignedat}
\end{equation}
where $\theta^1,\cdots,\theta^N$ are generated samples from the posterior distribution $p(\theta|\mathcal{D})$. Efficient sampling of the posterior distribution $p(\theta|\mathcal{D})$ is therefore important in the BNN framework, and will be discussed in Section \ref{sec:sampling}. The overall procedure for solving forward problems using BNN is described in Algorithm \ref{alg:BNN_method}. The choice of an appropriate NN model (Algorithm \ref{alg:BNN_method}, line 1) is considered in Section \ref{sec:FFN}.

\begin{algorithm}[h]
    \caption{BNN method to solve the forward problem of stochastic partial differential equations}
    \textbf{Input:} Data $\mathcal{D} = \{[\mathbf{f}(\omega_j),\mathbf{g}(\omega_j),\mathbf{k}(\omega_j)]\}_{j=1}^{N_\omega}$, prior distribution $p(\theta)$, new input $x^*$
    \begin{algorithmic}[1]
    \State Define an NN model with outputs $K(x,\theta)$ and $U(x,\theta)$ respectively for the parameter $k(x,\omega)$ and solution $u(x,\omega)$ of (\ref{eqn:BVP}).
    \State Estimate the density $p_{\mathbf{f},\mathbf{g},\mathbf{k}}$ using the data $\mathcal{D}$ as in (\ref{eqn:gmm}).
    \State Generate samples $\theta^1,\cdots,\theta^N$ from the posterior distribution (\ref{eqn:Bayes_rule}) using HMC in Algorithm \ref{alg:HMC}.
    \State Compute $U(x^*,\theta^i)$ for each $\theta^i$.
    \State Estimate statistics such as mean and variance as in (\ref{eqn:mean_and_variance_estimate}).
    \end{algorithmic}
    \textbf{Output:} Estimated statistics of the solution $u$ for the new input $x^*$.
    \label{alg:BNN_method}
\end{algorithm}

For inverse problems, we form the likelihood distribution with additional data of $u$:
\begin{equation*}
p(\mathcal{D}|\theta) = p_{\mathbf{f},\mathbf{g},\mathbf{u}}([\mathbf{F}(\theta),\mathbf{G}(\theta),\mathbf{U}(\theta)]),
\end{equation*}
where
\begin{equation*}
\mathbf{U}(\theta) = [U(x^{u,1},\theta),\cdots,U(x^{u,N_u},\theta)],
\end{equation*}
and $p_{\mathbf{f},\mathbf{g},\mathbf{u}}$ is a joint PDF of $\mathbf{f},\mathbf{g}$, and $\mathbf{u}$. We can derive the posterior distribution using Bayes' rule (\ref{eqn:Bayes_rule}). The mean and variance of the parameter $k$ for new input $x^*$ can be estimated by a Monte Carlo estimator
\begin{align*}
\mathbb{E}K(x^*) &= \int K(x^*,\theta)p(\theta |\mathcal{D})d\theta \approx \frac{1}{N} \sum_{i=1}^N K(x^*,\theta^i), \\
\text{Var}K(x^*) &= \int (K(x^*,\theta)-\mathbb{E}U^*)^2p(\theta |\mathcal{D})d\theta \approx \frac{1}{N} \sum_{i=1}^N (K(x^*,\theta^i) - \mathbb{E}K^*)^2,
\end{align*}
where $\theta^1,\cdots,\theta^N$ are generated samples from the posterior distribution. The algorithm for solving inverse problems is similar to Algorithm \ref{alg:BNN_method}.

\subsection{Posterior sampling with Hamiltonian Monte Carlo}\label{sec:sampling}

A major obstacle when using Bayes' rule (\ref{eqn:Bayes_rule}) is that high-dimensional integrals in the denominators are computationally intractable. In the BNN framework, two methods are commonly used to generate samples from the unnormalized posterior form: variational inference \cite{graves2011practical,blundell2015weight,gal2016dropout} and Markov chain Monte Carlo (MCMC) \cite{welling2011bayesian,neal2012bayesian}. 

Variational inference builds an easy-to-sample distribution model $q_\phi(z)$ with the parameter $\phi$ that approximates the target distribution of $z$ of which the unnormalized form is denoted by $p(z)$. To be precise, the purpose of variational inference is to find $\phi$ that minimizes the Kullback-Leibler divergence of two distributions $q_\phi(z)$ and $p(z)$,
\begin{equation*}
\text{KL}(q_\phi(z)\|p(z)/Z) = \int q_\phi(z)\log \frac{q_\phi(z)}{p(z)/Z} = \int q_\phi(z)(\log q_\phi(z) - \log p(z)) + \log Z,
\end{equation*}
where $Z$ is a normalizing constant of $p(z)$, i.e., $Z=\int p(z)dz$. The last term does not depend on $\phi$, so finding $\phi$ that minimizes the first term is sufficient. Therefore, we do not need to compute the normalizing constant $Z$ for the sampling. Once we find $\phi$, we easily obtain samples of $z$ using $q_\phi(z)$ instead of $p(z)$. However, the selection of an appropriate approximate model $q_\phi(z)$ for the posterior of NN parameters is not an easy task.

MCMC creates a Markov chain for which the stationary distribution is the target distribution, which means the elements of the chain can be regarded as samples from the target distribution that can be used for Monte Carlo estimation. The Metropolis-Hastings algorithm is commonly used to generate the Markov chain. To be precise, the algorithm proposes the next element $z'$ from the current element $z$ by exploiting the proposal density $g(z'|z)$, and accepts the proposed sample with the probability
\begin{equation*}
\alpha = \min(1,\frac{p(z')g(z|z')}{p(z)g(z'|z)}).
\end{equation*}
We only use the ratio of $p(z)$ when we compute the acceptance rate $\alpha$, so normalization of the target distribution is not necessary. However, a proposal density with a low acceptance rate may require an excessive number of samples for a decent approximation, and this possibility can reduce the efficiency of MCMC.

Hamiltonian Monte Carlo (HMC, Algorithm \ref{alg:HMC}) is an algorithm of MCMC, for which proposal density provides a high acceptance rate even for a high-dimensional state space. This trait is particularly important for BNNs with high-dimensional state spaces due to numerous NN parameters. HMC uses Hamiltonian dynamics to propose the next sample. A particle with momentum searches for a new sample while exploring an energy field constructed by the unnormalized form of the target distribution.

To find samples from density $p(z)$, HMC introduces an auxiliary momentum $r$ with distribution $p(r)$ that depends only on the size of momentum $|r|$ and not on $z$. With the help of auxiliary momentum $r$, HMC proposes a sample by the following steps: (i) obtain a new sample $r$ from $p(r)$, (ii) explore a new sample $z'$ (and $r'$) by solving Hamiltonian equation
\begin{equation*}
    \dot{z}=\nabla_r H,\;\;\dot{r}=-\nabla_z H
\end{equation*}
with a Hamiltonian $H = -\log p(z) -\log p(r) =: V(z) + K(r)$ for some time interval. This process means that the proposed sample is the destination of a particle moving on energy field $V(z)$ for a certain time period. The proposed sample generated in this Markov chain is always accepted due to the Hamiltonian conservation and time-reversal symmetry of the dynamics.

The Hamiltonian equation has no exact closed-form solution, so the leapfrog scheme (Algorithm \ref{alg:HMC}, lines 5 to 7) is commonly used. The leapfrog scheme keeps the time-reversal symmetry, but no longer preserve the Hamiltonian exactly. Therefore, the acceptance rate
\begin{equation}\label{eqn:HMC_acceptance}
\alpha = \min(1,\frac{p(z')g(z|z')}{p(z)g(z'|z)}) = \min (1, \frac{p(z')p(-r')}{p(z)p(r)}) = \min (1, \exp(H(z,r)-H(z',r')))
\end{equation}
is introduced to ensure the convergence of the Markov chain. To explain, $g(z'|z)$ is the probability density of $z'$ being the next state of $z$, which is equal to the probability density of $r$ being the initial momentum at the $z$ state; therefore, $g(z'|z) = p(r)$. Conversely, $g(z|z')$ is equal to the probability density that $-r'$ is the initial momentum at the $z'$ state; a particle arrives in the previous state of $z'$ when it moves from $z'$ with the initial momentum $-r'$ by the time-reversal symmetry. Because $p(r)$ only depends on $|r|$, $p(-r')=p(r')$, and hence we derive the acceptance rate in (\ref{eqn:HMC_acceptance}). A small leapfrog step size increases the acceptance rate by reducing the difference between Hamiltonians of the current and next states but at the expense of computational cost. High acceptance rate must be traded off with computational cost, so the hyperparameter of the scheme must be chosen appropriately for each situation. The overall procedure of HMC is summarized in Algorithm \ref{alg:HMC}.

The computational cost of obtaining one HMC sample is mainly determined by the number of leapfrog steps $M$ and the computation of $\nabla_z V$ (Algorithm \ref{alg:HMC}, lines 5 and 7) in each leapfrog step. It has become suggested that the leapfrog step size $\delta$ should be scaled as $O(d^{-1/4})$ to maintain high acceptance rate as $d \rightarrow \infty$ where $d$ is the dimension of the state space \cite{beskos2013optimal,mangoubi2018dimensionally}. This implies that HMC requires $M=O(d^{1/4})$ steps to solve the Hamiltonian equation in the equal time interval. In addition, using automatic differentiation and sufficient parallel computation, the cost of computing the gradient $\nabla_z V$ barely increases as $d$ increases. These observations provide a strong motivation to apply HMC to high-dimensional problems. We demonstrate the efficacy of HMC in high-dimensional problems in Section \ref{sec:example}.

\begin{algorithm}[h]
    \caption{Hamiltonian Monte Carlo}
    \textbf{Input:} Initial sample $z^0$, number of burn-in step $L$, number of samples $N$, number of leapfrog steps $M$, leapfrog step size $\delta$, target distribution $p(z)$, distribution of auxiliary momentum $p(r)$
    \begin{algorithmic}[1]
    \ForAll{$i=0,\cdots,L+N-1$}
        \State $r^i \sim \pi(r)$
        \State $r^{(0)} = r^i, z^{(0)} = z^i$
        \ForAll{$j=0,\cdots,M-1$}
            \State $r^{(j+1/2)} = r^{(j)} - \frac{\delta}{2}\nabla_z V\vert_{z=z^{(j)}}$
            \State $z^{(j+1)} = z^{(j)} + \delta\nabla_r K\vert_{r=r^{(j+1/2)}}$
            \State $r^{(j+1)} = r^{(j+1/2)} - \frac{\delta}{2}\nabla_z V\vert_{z=z^{(j+1)}}$
        \EndFor
        \State $r' = r^{(M)}, z' = z^{(M)}$
        \State Accept $z'$ as a new sample with probability $\alpha=\min (1, \exp(H(z,r)-H(z',r')))$.
        \If {$z'$ is accepted}
            \State $z^{i+1} = z'$
        \Else
            \State $z^{i+1} = z^i$
        \EndIf
    \EndFor
    \end{algorithmic}
    \textbf{Output:} Samples $z^{L+1},\cdots,z^{L+N}$
    \label{alg:HMC}
\end{algorithm}

\subsection{Fourier feature network}\label{sec:FFN}

In this section, we review the Fourier feature network (FFN) \cite{tancik2020fourier}. FFN is a network that overcomes spectral bias \cite{rahaman2019spectral,cao2019towards,ronen2019convergence}; a commonly observed phenomenon of fully-connected networks (FCNs) that prevents them from learning high-frequency features. The spectral bias also exists in PINN tasks, so FFN outperforms the FCN when used as a PINN model to solve PDE problems containing high-frequency functions \cite{wang2021eigenvector}. We use FFN as a surrogate model for the solution of SPDE to deal with the source terms with various frequencies in Section \ref{sec:example}.

An FFN with input dimension $n$ is constructed using a Fourier feature embedding of the inputs
\begin{equation*}
\gamma(x) = \begin{bmatrix} \cos{Bx} \\ \sin{Bx} \end{bmatrix},
\end{equation*}
followed by an FCN. Each entry of a matrix $B \in \mathbb{R}^{b \times n}$ is generated from a normal distribution $\mathcal{N}(0,\sigma^2)$ for some positive hyperparameter $\sigma$. This simple variation of the FCN allows the network to effectively learn high-frequency features, and thereby increases its efficiency in many tasks such as image regression and computed tomography \cite{tancik2020fourier}.

Increase in hyperparameter $\sigma$ increases the frequency of features that an FFN can learn. In addition, using neural tangent kernel theory \cite{jacot2018neural,arora2019exact,lee2019wide}, a specific value of $\sigma$ induces fast convergence of a specific frequency \cite{wang2021eigenvector}. This observation means that we must choose an appropriate $\sigma$ depending on the frequencies included in the problem, and that excessively large $\sigma$ can lead to overfitting. Moreover, in a multiscale problem in which various frequencies appear, using FFN is no longer effective because it induces fast convergence only for the specific frequency.

Therefore, multiscale FFN is introduced by applying the multiple Fourier feature embeddings corresponding to different $\sigma$ to input, passing the embedded inputs to the same FCN, and finally passing the concatenation of all outputs to a single layer. The described forward pass of multiscale FFN $U(x)$ is summarized as
\begin{align*}
H_0^{(i)} &= \gamma^{(i)}(x) = \begin{bmatrix} \cos{B^{(i)}x} \\ \sin{B^{(i)}x} \end{bmatrix}\;\;\text{for $i=1,\cdots,I$},\\
H_{t+1}^{(i)} &= \phi(W_t H_t^{(i)} + b_t)\;\;\text{for $i=1,\cdots,I$, $t=0,\cdots,T-1$}, \\
U(x) &= W_T\begin{bmatrix} H_T^{(1)} \\ \vdots \\ H_T^{(I)} \end{bmatrix} + b_T,
\end{align*}
where $\gamma^{(i)}$ are multiple Fourier feature embeddings, $\phi$ is an activation function, $W_t$ are weights of the FCN, and $b_t$ are its biases. Each entry of $B^{(i)}$ is generated from a normal distribution $\mathcal{N}(0,\sigma_i^2)$ for different $\sigma_i$. Empirical results have shown that multiscale FFNs induce fast convergence for various frequencies, and yield better results than simple FFNs in multiscale problems \cite{wang2021eigenvector}.

\section{Numerical examples}\label{sec:example}

In this section, we demonstrate the efficacy of our proposed method by applying it to numerical examples. We solve three types of SPDE problems to demonstrate that our method works for both forward and inverse problems, and that it provides good prediction even for high-dimensional problems.

In each example, the prior distribution of the parameter is set to independent standard normal distribution. In the HMC setup, the distribution of the auxiliary momentum $r$ is a standard normal distribution, the number of burn-in steps is $L=\text{1,000}$, and the number of BNN samples is $N=\text{4,000}$. To evaluate the BNN samples, we compare their sample mean and standard deviation (STD) to the reference values. We obtain the reference solution by using the Monte Carlo method with 500,000 samples for one-dimensional (1D) problems and 100,000 samples for two-dimensional (2D) problems. We use the FEniCS library \cite{logg2012automated,alnaes2015fenics} to solve the deterministic PDE that corresponds to each Monte Carlo sample. The number of training snapshot data on the spatial sensors is $N_\omega=\text{20,000}$.

\subsection{Stochastic process}\label{sec:process}

We first test our method to generate samples of a random process. Although this is not an SPDE solving problem, the framework is the same because our problem setup is reduced to
\begin{equation*}
u(x,\omega) = f(x,\omega),\;\;\text{for $x\in D=[-1,1],\omega \in \Omega$}.
\end{equation*}
We generate BNN samples for the random process with the distribution
\begin{equation*}
\log (f(x) - 0.5) \sim \mathcal{GP}(\sin \pi x, \sigma_f^2 \exp \Big(-\frac{|x-x'|^2}{2l_f^2}\Big))\;\;\text{for $x \in D$},
\end{equation*}
where $\sigma_f=0.1$ and $l_f=0.1$. We assume $N_f=41$ equidistant sensors of $f$ on the domain $D$. An NN model $U$ is a multiscale FFN with two Fourier feature embeddings
\begin{equation*}
\gamma^{(1)}(x) = \begin{bmatrix} \cos{B^{(1)}x} \\ \sin{B^{(1)}x} \end{bmatrix},\;\;\gamma^{(2)}(x) = \begin{bmatrix} \cos{B^{(2)}x} \\ \sin{B^{(2)}x} \end{bmatrix},
\end{equation*}
where $B^{(i)} \in \mathbb{R}^{7 \times 1}$ and each entry of $B^{(i)}$ are generated from a normal distribution $\mathcal{N}(0,\sigma_i^2)$ with $\sigma_1=1.0$ and $\sigma_2=5.0$. The embedded inputs pass through an FCN that has sine activation and that consists of one hidden layer with 200 units. In the HMC setup, the number of leapfrog steps is $M=100$, and the time step is $\delta=1 \times 10^{-3}$.

\begin{figure}[htbp]
    \centering
    \begin{subfigure}{0.32\textwidth}
        \centering
        \includegraphics[height=4.5cm]{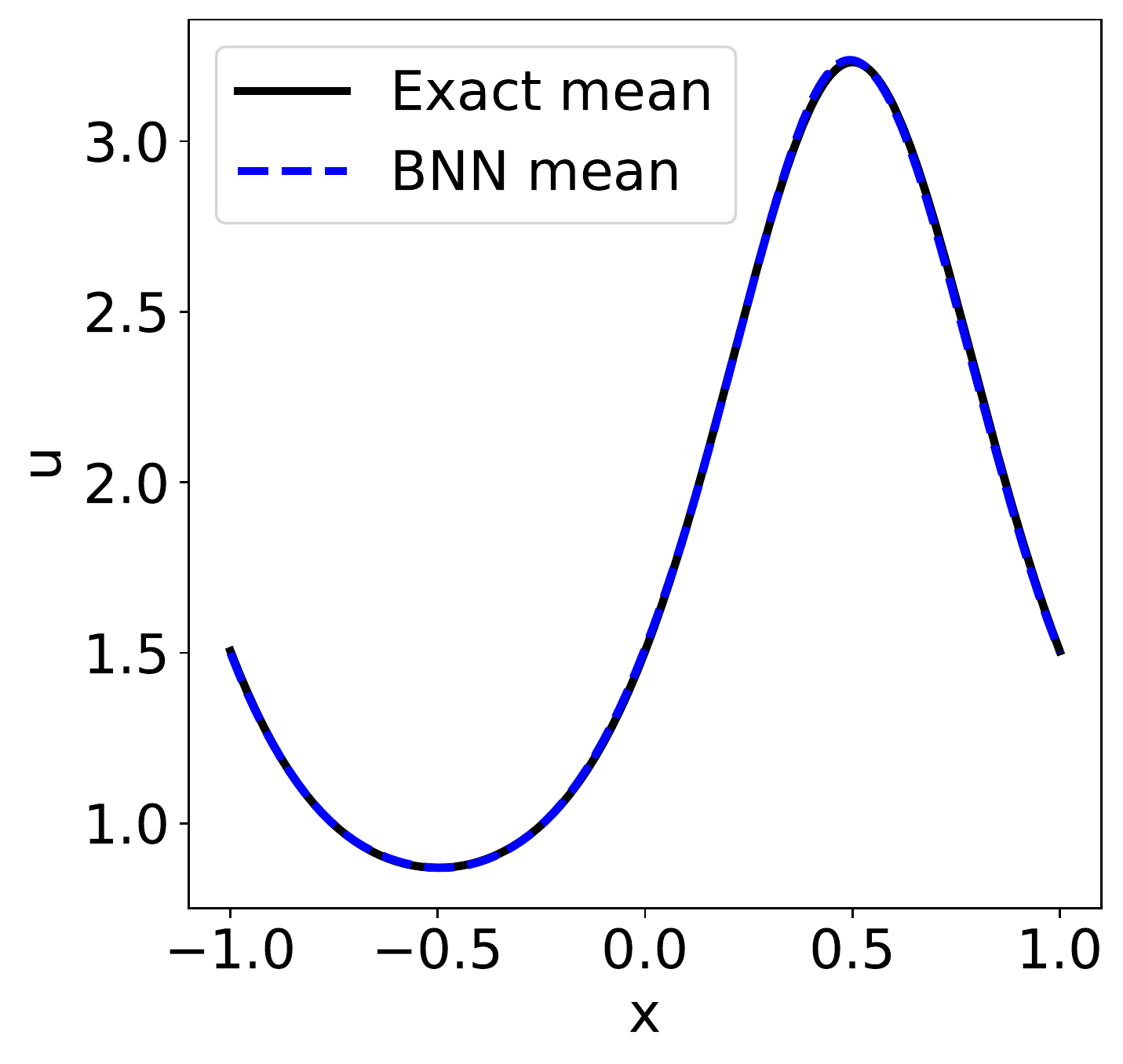}
        \caption{}
    \end{subfigure}
    \begin{subfigure}{0.32\textwidth}
        \centering
        \includegraphics[height=4.5cm]{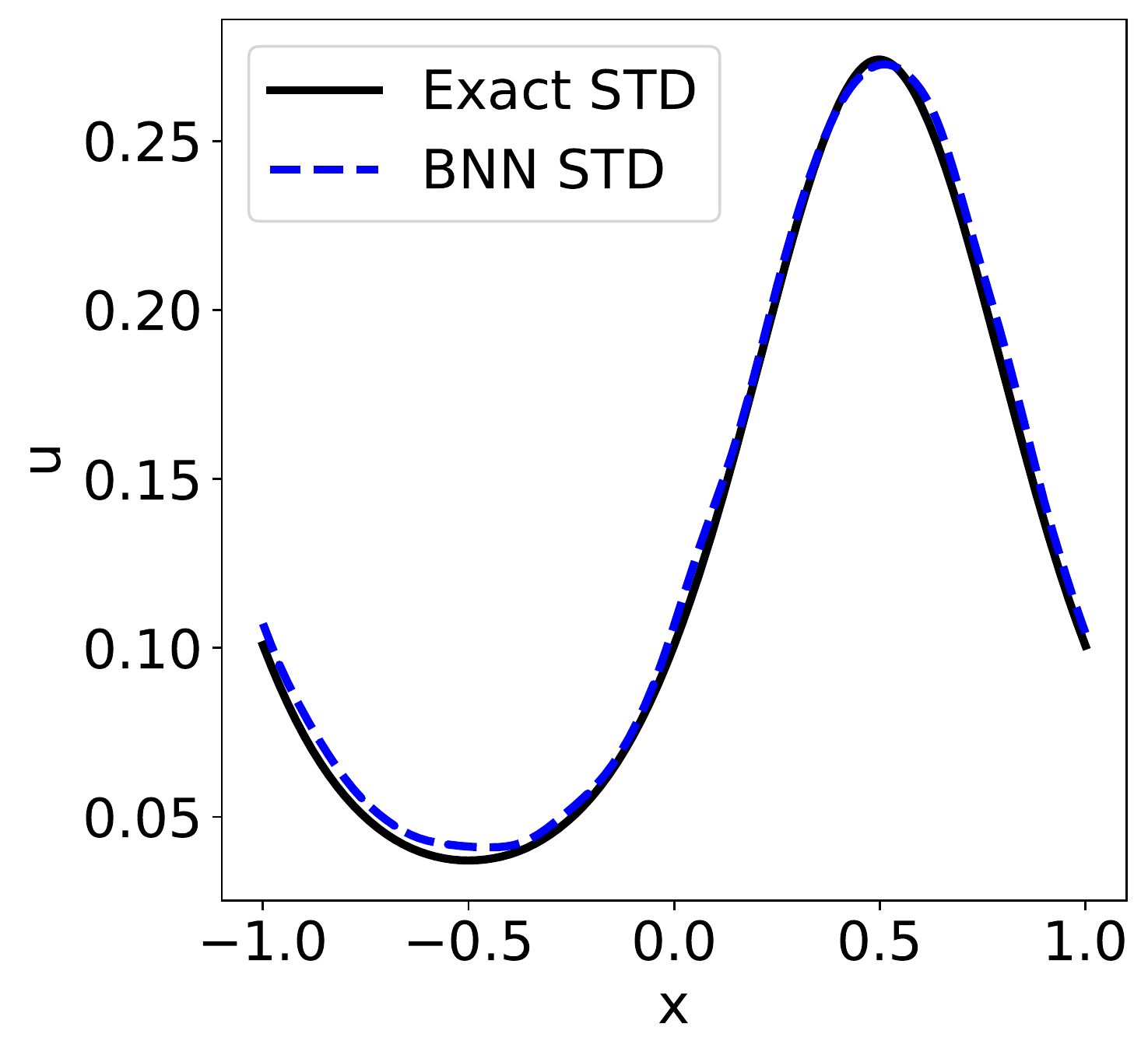}
        \caption{}
    \end{subfigure}
    \begin{subfigure}{0.32\textwidth}
        \centering
        \includegraphics[height=4.5cm]{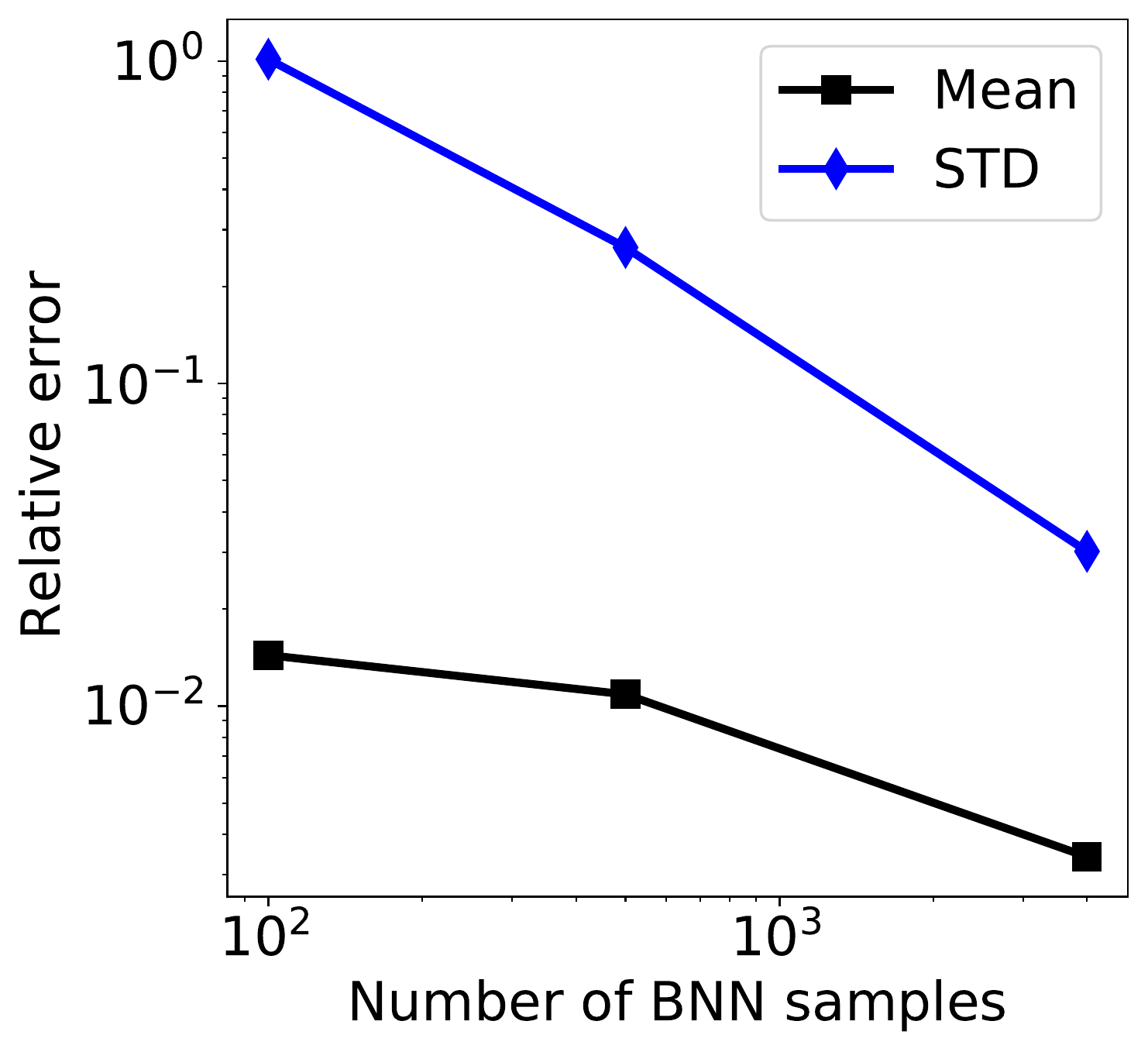}
        \caption{}
    \end{subfigure}

    \caption{Results of sampling stochastic process using BNN. Predicted (a) mean and (b) STD of the samples. (c) Relative error of mean and STD vs number of BNN samples.}
   \label{fig:process_error}
    
\end{figure}

\begin{figure}[htbp]
    \centering
    \begin{subfigure}{0.32\textwidth}
        \centering
        \includegraphics[height=4.3cm]{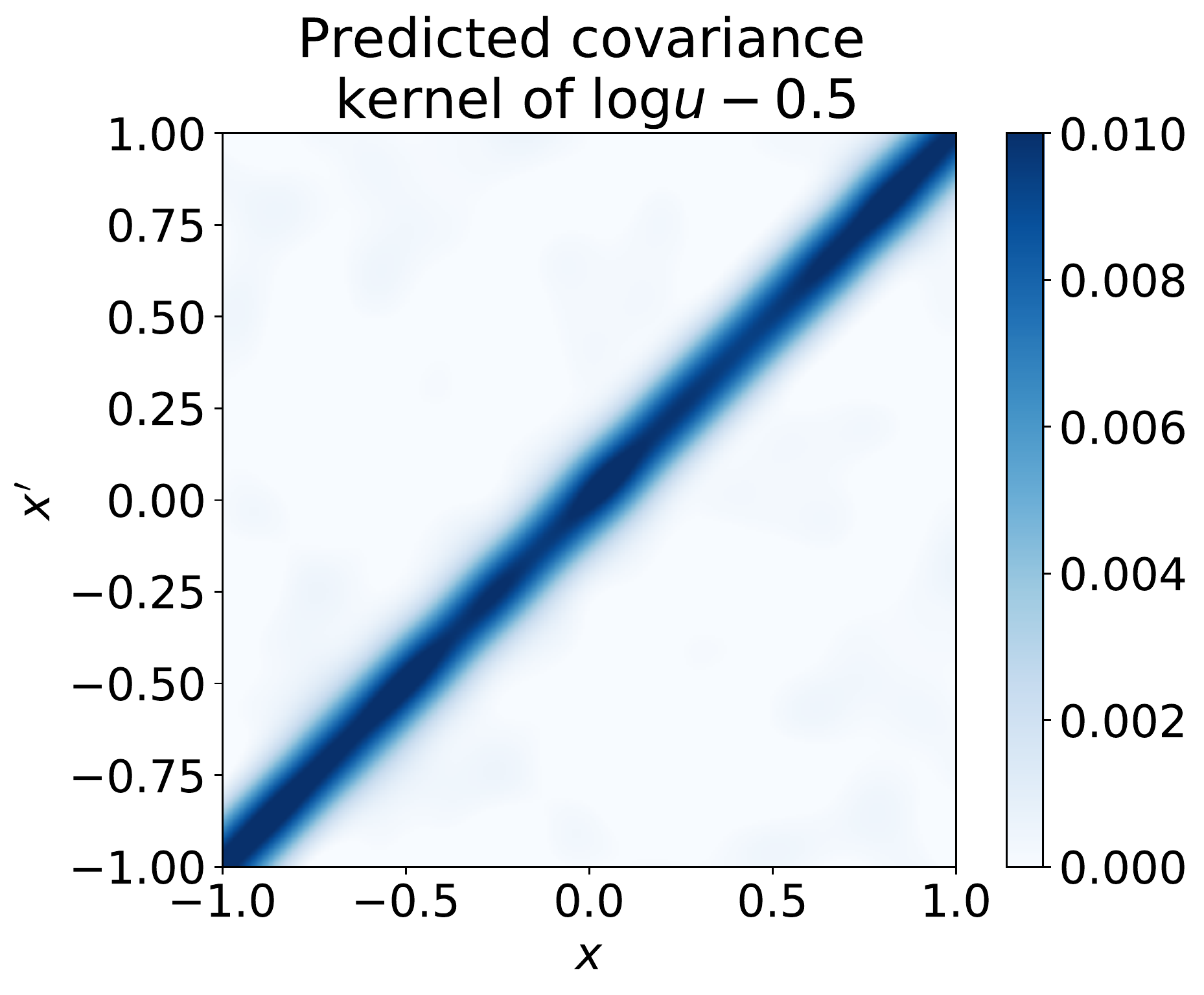}
        \caption{}
    \end{subfigure}
    \begin{subfigure}{0.32\textwidth}
        \centering
        \includegraphics[height=4.3cm]{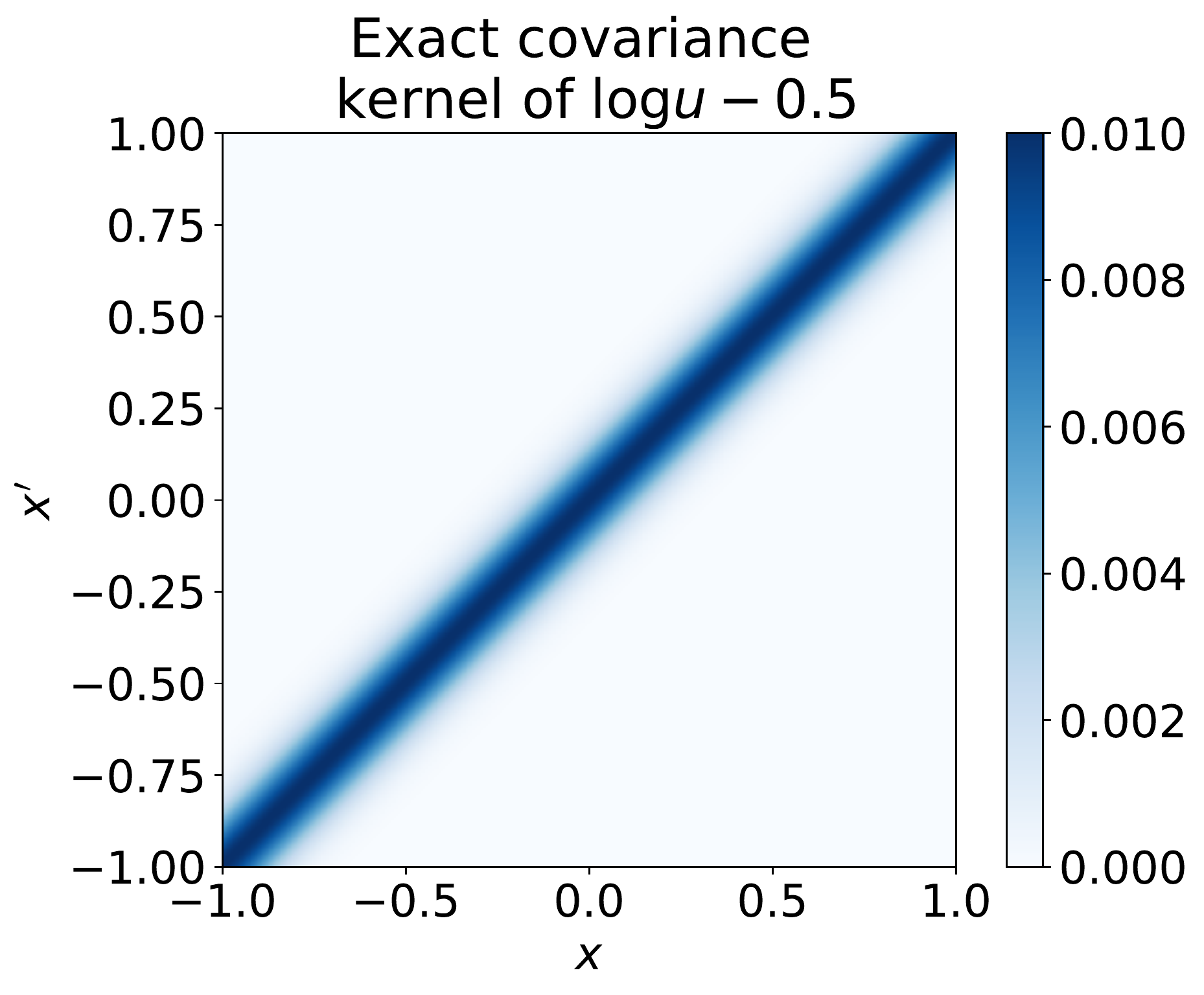}
        \caption{}
    \end{subfigure}
    \begin{subfigure}{0.32\textwidth}
        \centering
        \includegraphics[height=4.3cm]{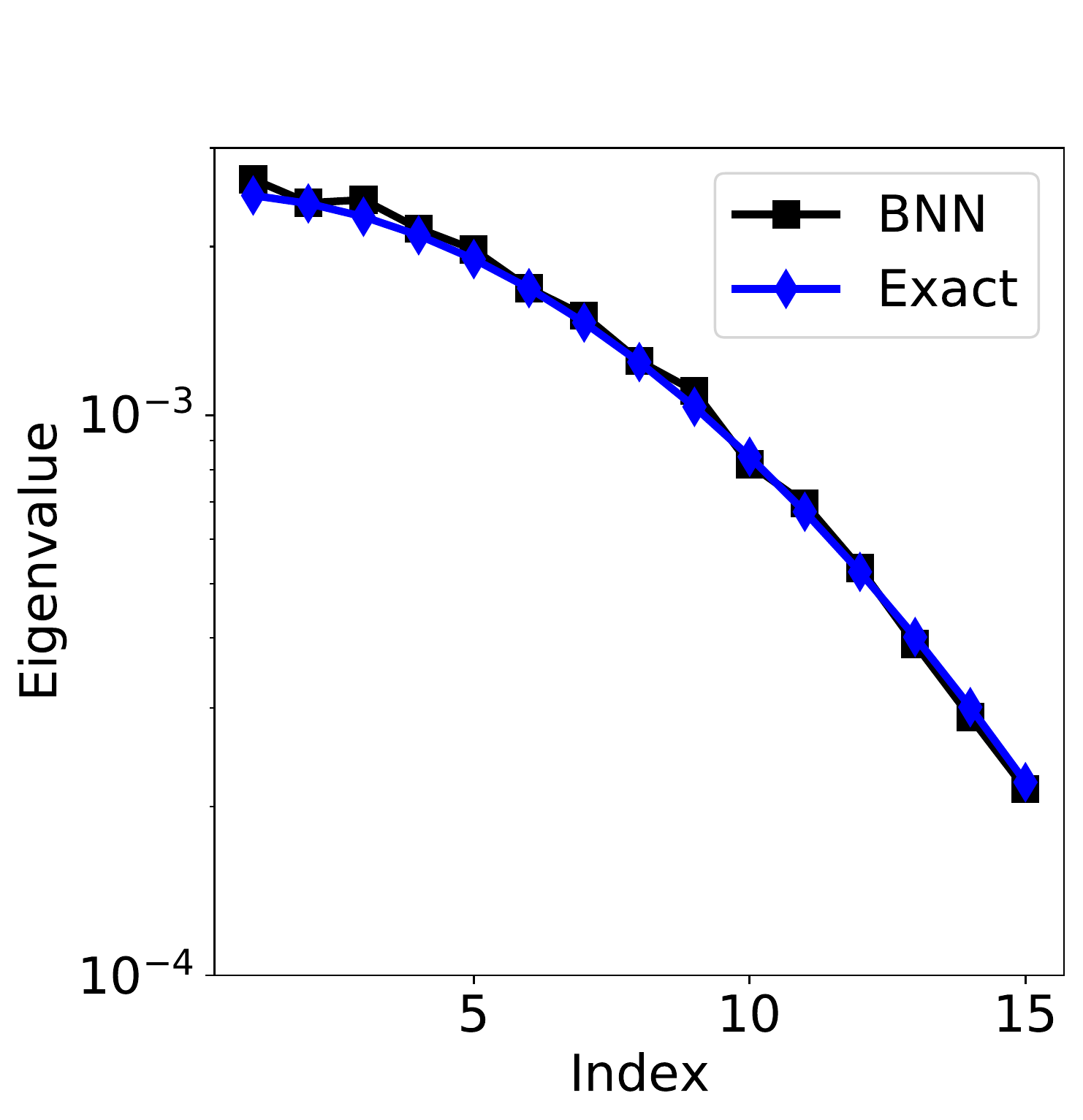}
        \caption{}
    \end{subfigure}

    \caption{(a) Predicted covariance kernel and (b) exact covariance kernel of $\log u - 0.5$ where $u$ is the given stochastic process. (c) Eigenvalues of the predicted and exact covariance kernels.}
   \label{fig:process_kernel}
    
\end{figure}

The mean and STD of our BNN samples provide good approximations and their error decreases as the number of BNN samples increases (Figure \ref{fig:process_error}); i.e., the Monte Carlo approximation converges. Furthermore, the computed covariance kernel of $\log u -0.5$ and its eigenvalues agree well with those of the exact kernel (Figure \ref{fig:process_kernel}). Using the information obtained from scattered sensors, our method predicts the random process well throughout the entire domain.

\subsection{The forward problem of 1D Poisson equation}\label{sec:1dpoisson}

In this section, we solve the following stochastic linear Poisson equation on a 1D domain:
\begin{align*}
-\frac{d^2}{dx^2}u(x,\omega) &= f(x,\omega)\;\;\text{for $x \in D=[-1,1]$}, \\
u(-1),u(1) &= 0.
\end{align*}
The source term $f$ is set to Gaussian process with the Mat\'{e}rn kernel:
\begin{equation}\label{eqn:GP}
f(x) \sim \mathcal{GP}(10\sin \pi x, \sigma_f^2 \Big(1+\frac{\sqrt{5}|x-x'|}{l_f} + \frac{5|x-x'|^2}{3l_f^2}\Big) \exp \Big(-\frac{|x-x'|}{l_f}\Big))\;\;\text{for $x \in D$}
\end{equation}
where $\sigma_f=1.0$ and $l_f=0.1$. We consider $N_f=41$ equidistant sensors of $f$ on $D$ and two sensors of $u$ at two boundaries. We assume that the measurements of $u$ have noises that follow a normal distribution $\mathcal{N}(0,0.01^2)$, which makes sense physically and allows use of Bayes' rule. We use the same multiscale FFN as in Section \ref{sec:process}, but with $B^{(i)} \in \mathbb{R}^{10 \times 1},\sigma_1=1.0$, and $\sigma_2=7.0$. The number of leapfrog steps is $M=100$, and the time step is $\delta=1 \times 10^{-4}$.

\begin{figure}[htbp]
    \centering
    \begin{subfigure}{0.32\textwidth}
        \centering
        \includegraphics[height=4.5cm]{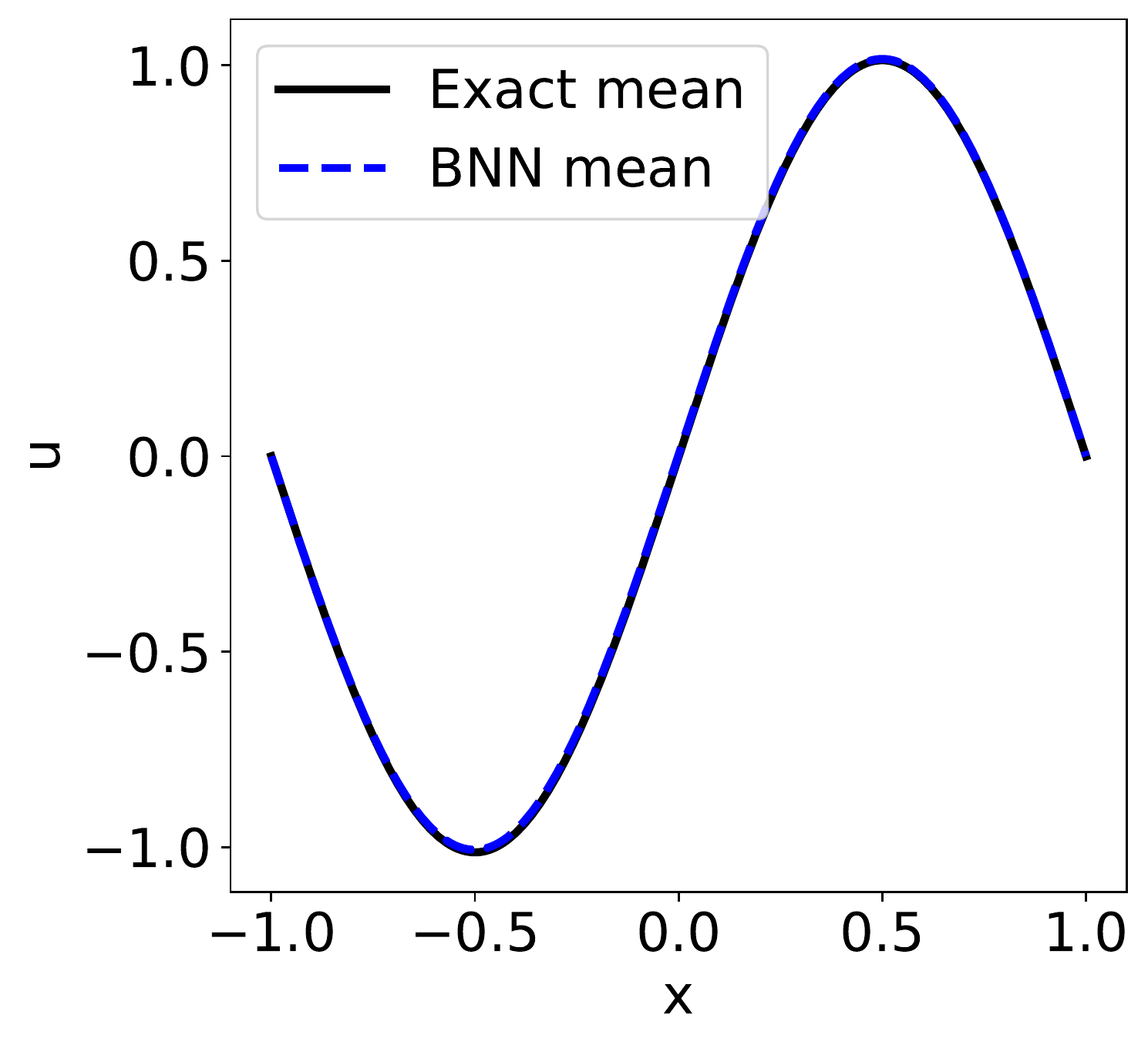}
        \caption{}
    \end{subfigure}
    \begin{subfigure}{0.32\textwidth}
        \centering
        \includegraphics[height=4.5cm]{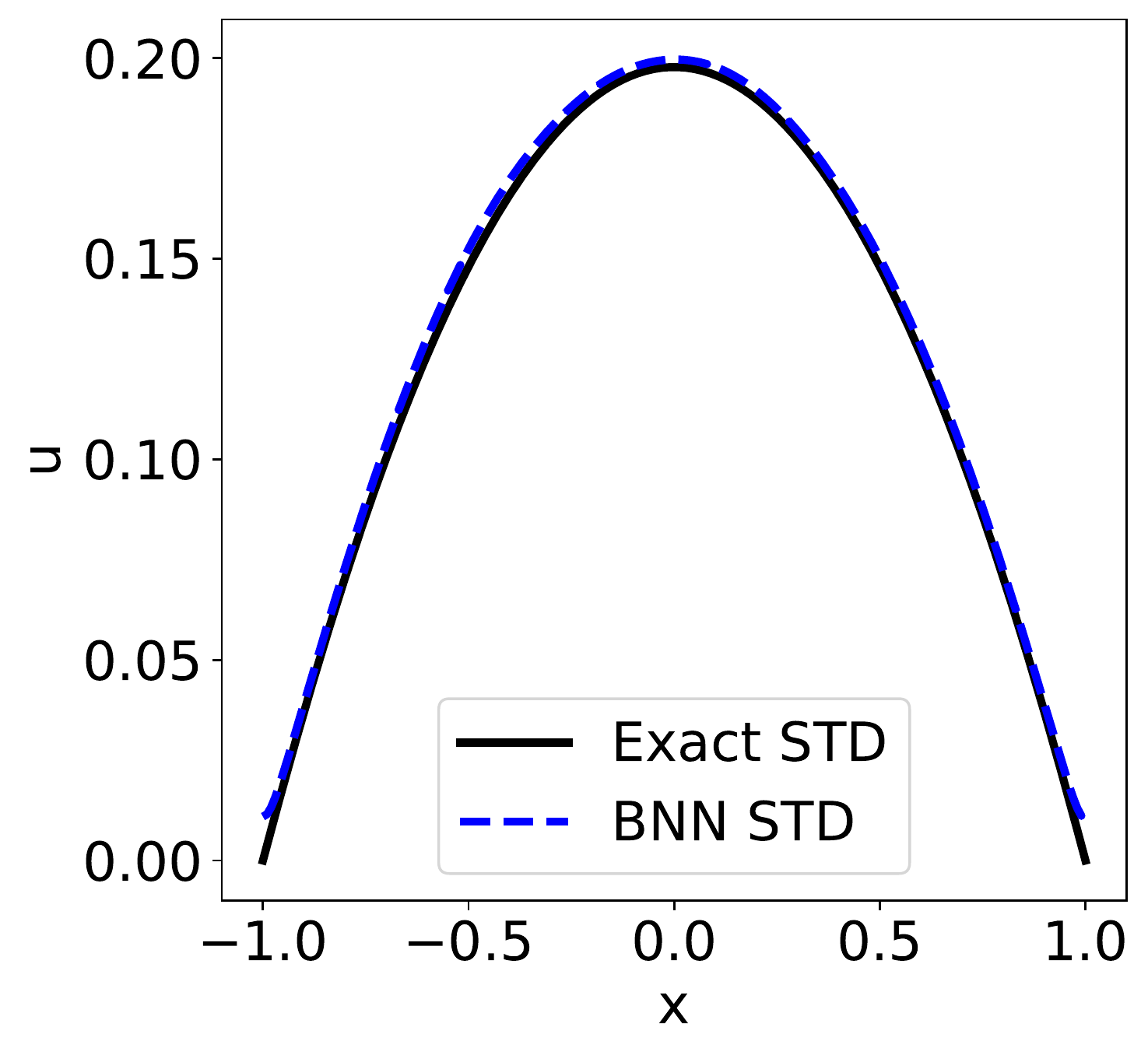}
        \caption{}
    \end{subfigure}
    \begin{subfigure}{0.32\textwidth}
        \centering
        \includegraphics[height=4.5cm]{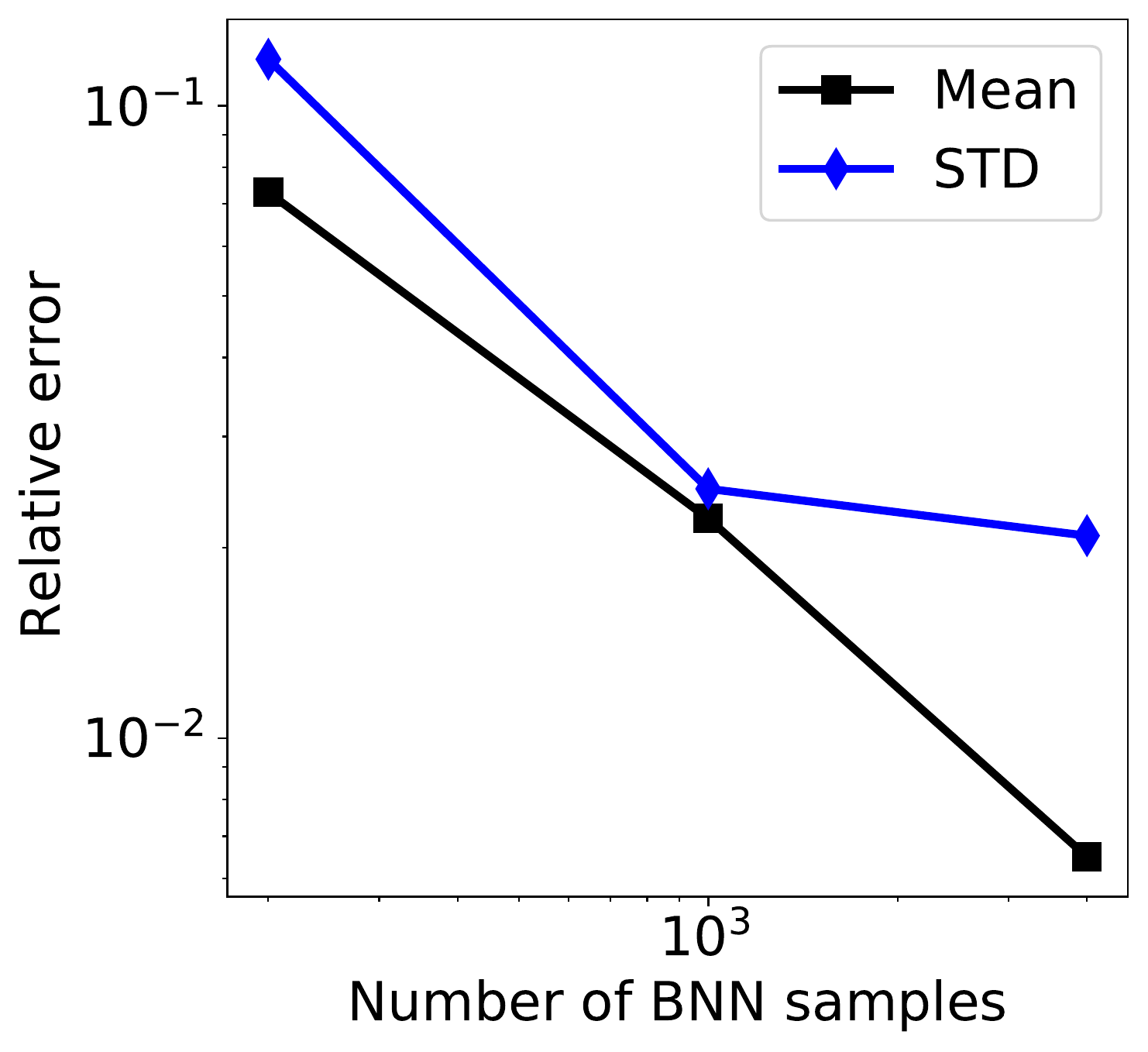}
        \caption{}
    \end{subfigure}
    \begin{subfigure}{0.40\textwidth}
        \centering
        \includegraphics[height=5cm]{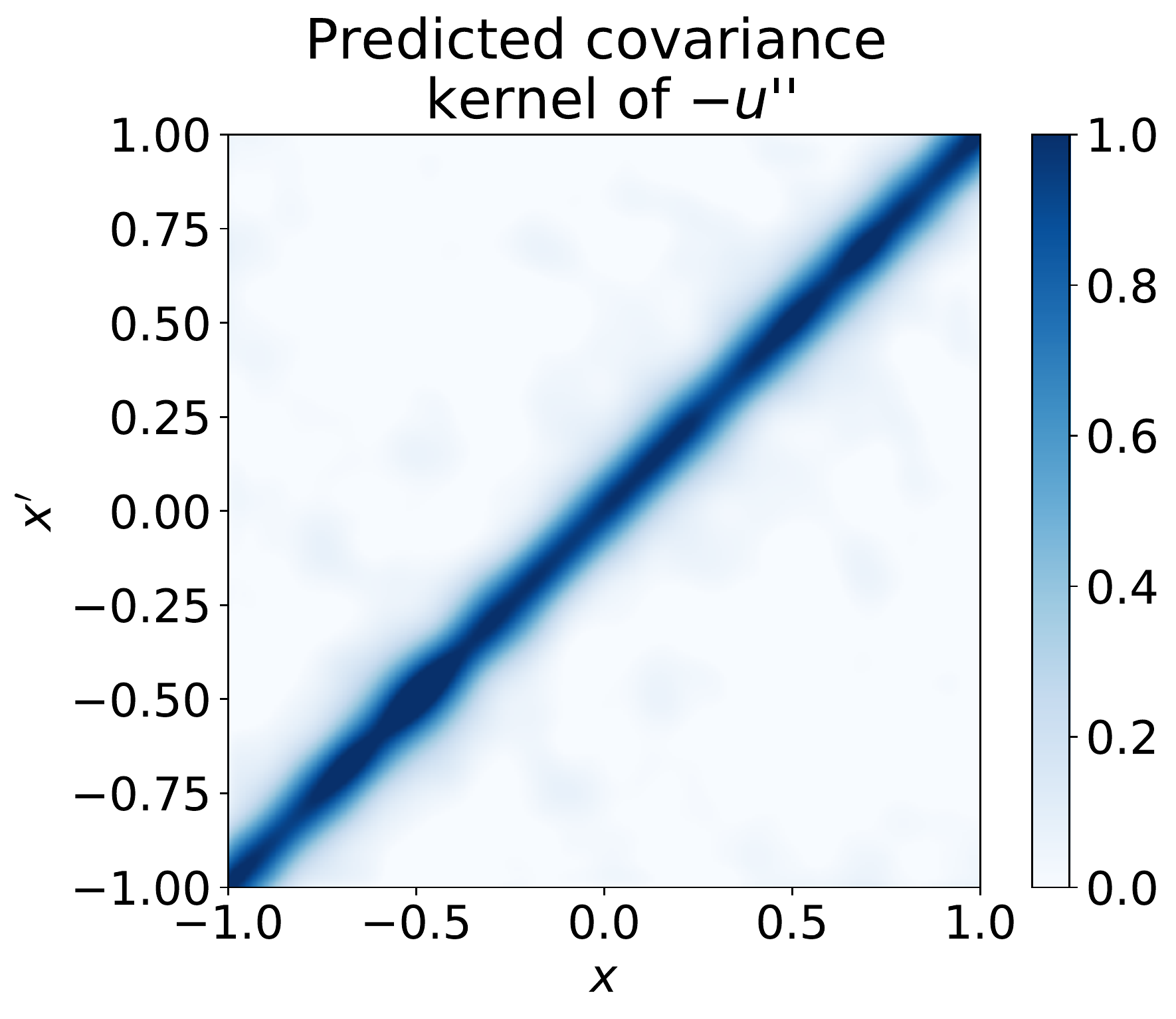}
        \caption{}
    \end{subfigure}
    \begin{subfigure}{0.40\textwidth}
        \centering
        \includegraphics[height=5cm]{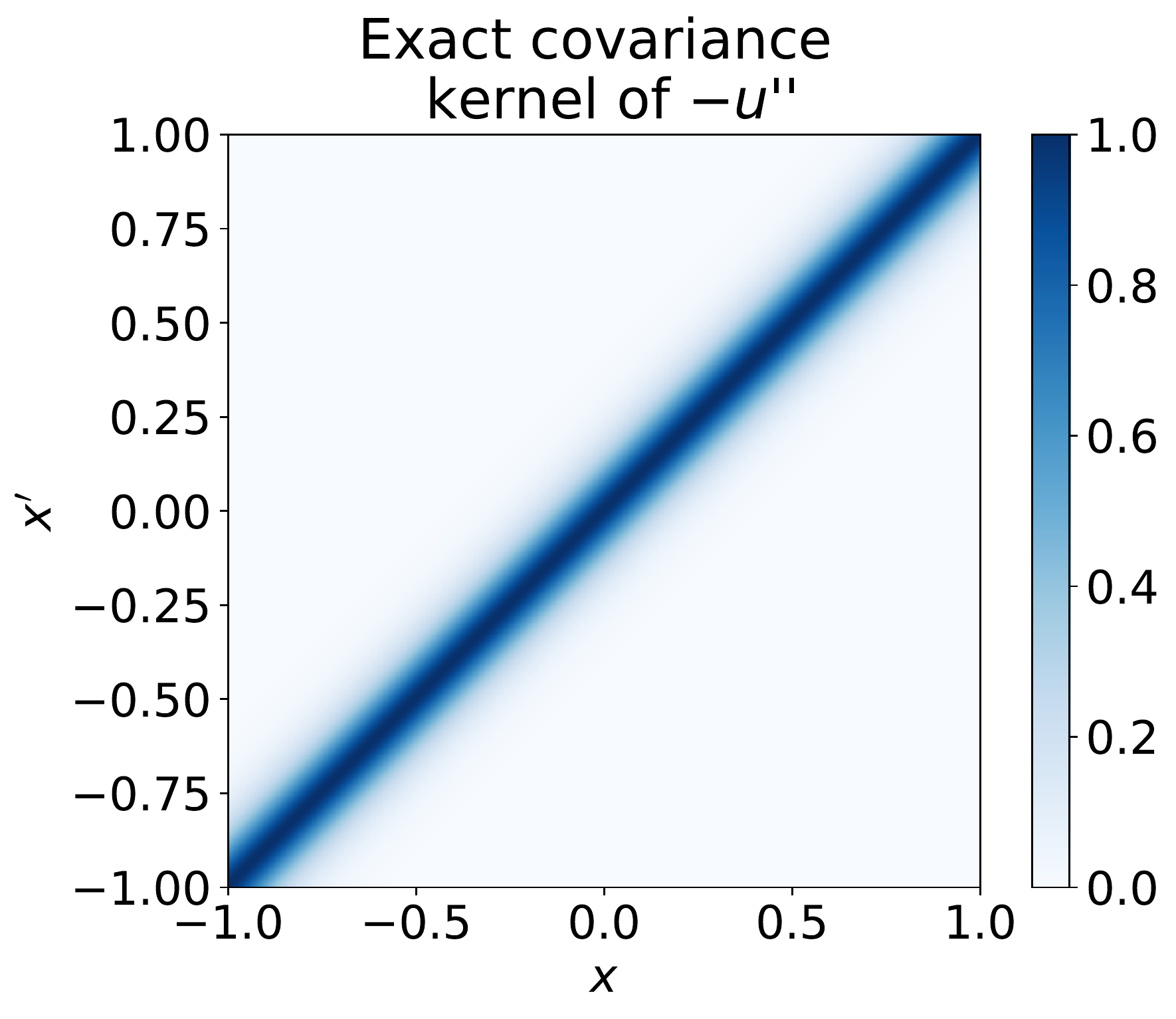}
        \caption{}
    \end{subfigure}

    \caption{Results of sampling the solution of Poisson equation by using a BNN where $\sigma_f=1.0$ and $l_f=0.1$. Predicted (a) mean and (b) STD of the solution. At the boundary, predicted STD is close to 0.01, which corresponds to the assumed noise STD. (c) Relative error of mean and STD vs number of BNN samples. (d) Predicted covariance kernel and (e) exact covariance kernel of $-u''$.}
   \label{fig:poisson_low}
    
\end{figure}

The mean of the computed solution samples gives a good approximation (Figure \ref{fig:poisson_low}). The solution samples also have a similar STD to the reference, but not at the boundary because of the measurement noise. The predicted STD at the boundary is approximately 0.01, which corresponds to the assumed noise STD. The covariance of the negative second derivative of $u$ shows good agreement with the prescribed kernel.

We next solve the equation when the correlation length is $l_f=0.03$. The correlation length of the kernel determines the dimension of random space required to approximate the random process. Table \ref{tab:dimension} shows the dimensions required to capture 99\% of the energy of Gaussian processes associated with the eigenvalues of the covariance kernel. A small correlation length induces a high dimension of the SPDE problem, which causes the so-called curse of dimensionality for traditional uncertainty quantification methods such as polynomial chaos \cite{xiu2002wiener}. 

\begin{table}[htbp]
    \centering
    \begin{tabular}{|c|c|c|c|c|c|}
    \noalign{\smallskip}\noalign{\smallskip}\hline
    Correlation length & $1.0$ & $0.3$ & $0.2$ & $0.1$ & $0.03$ \\
    \hline
    Dimension & 4 & 10 & 14 & 27 & 87 \\
    \hline
    \end{tabular}
    \caption{Dimension required to capture 99\% of the energy of Gaussian processes (\ref{eqn:GP}) associated with the eigenvalues of the covariance kernels for each correlation length.}
    \label{tab:dimension}
\end{table}

\begin{figure}[htbp]
    \centering
    \begin{subfigure}{0.32\textwidth}
        \centering
        \includegraphics[height=4.5cm]{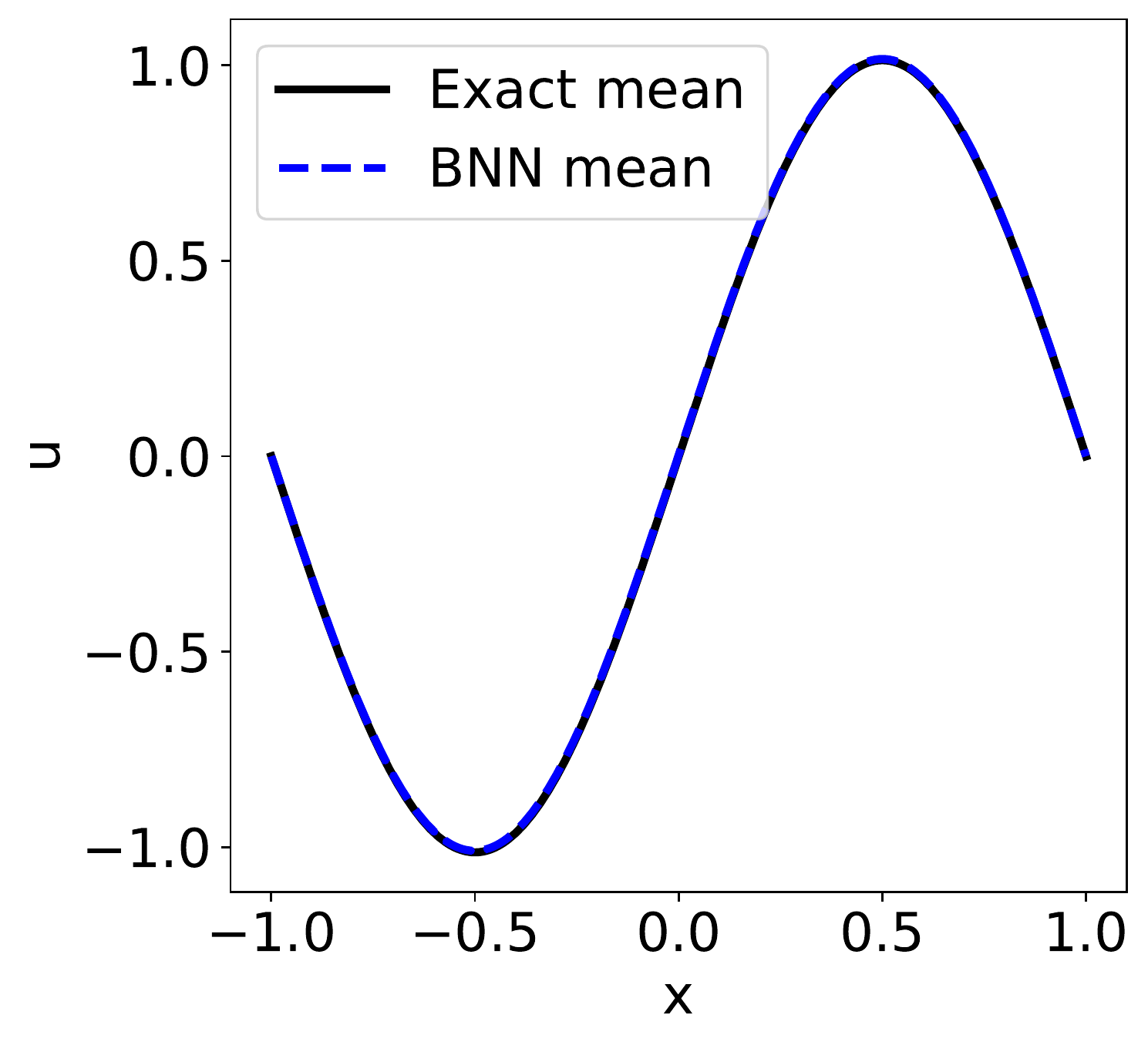}
        \caption{}
    \end{subfigure}
    \begin{subfigure}{0.32\textwidth}
        \centering
        \includegraphics[height=4.5cm]{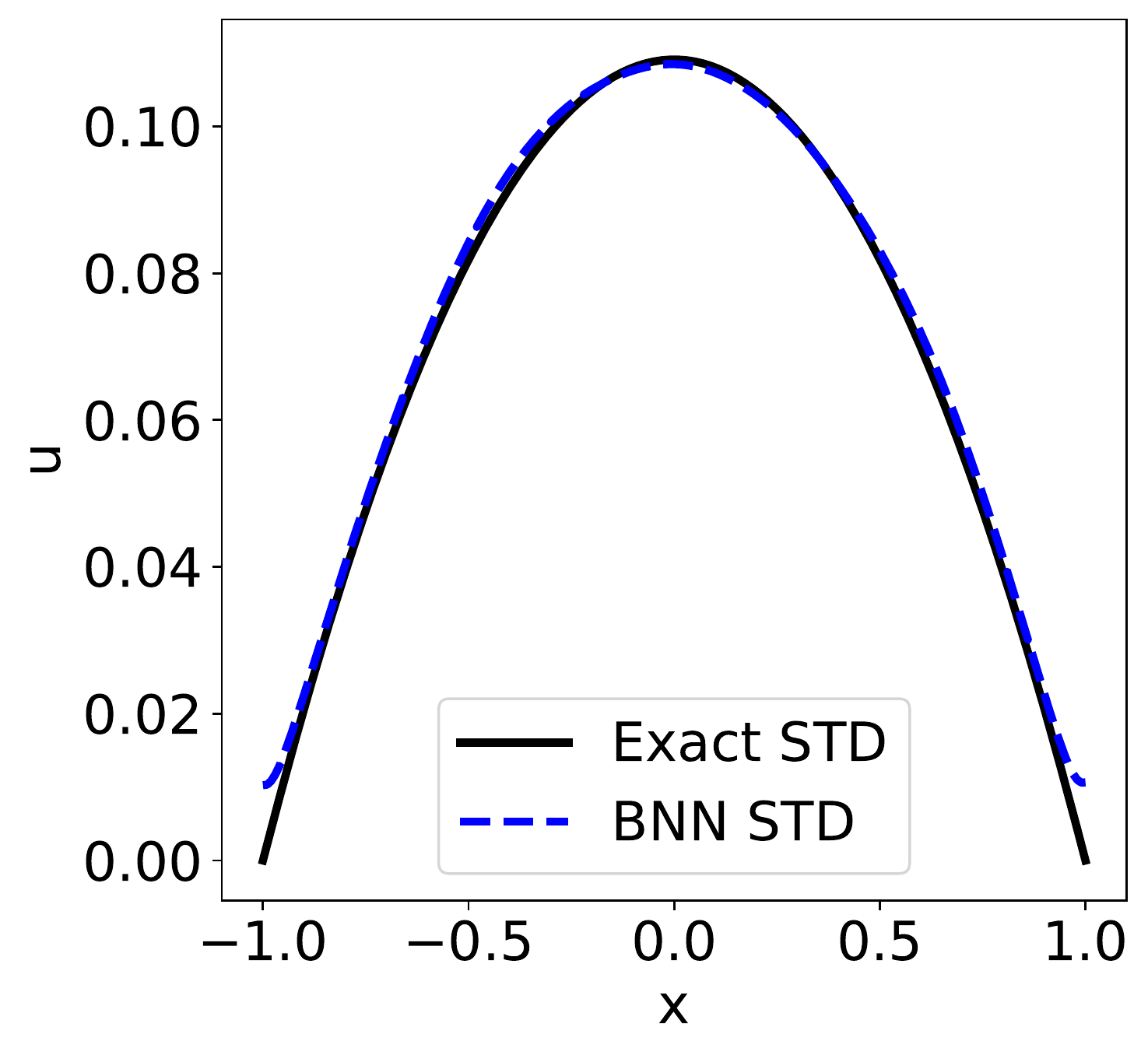}
        \caption{}
    \end{subfigure}
    \begin{subfigure}{0.32\textwidth}
        \centering
        \includegraphics[height=4.5cm]{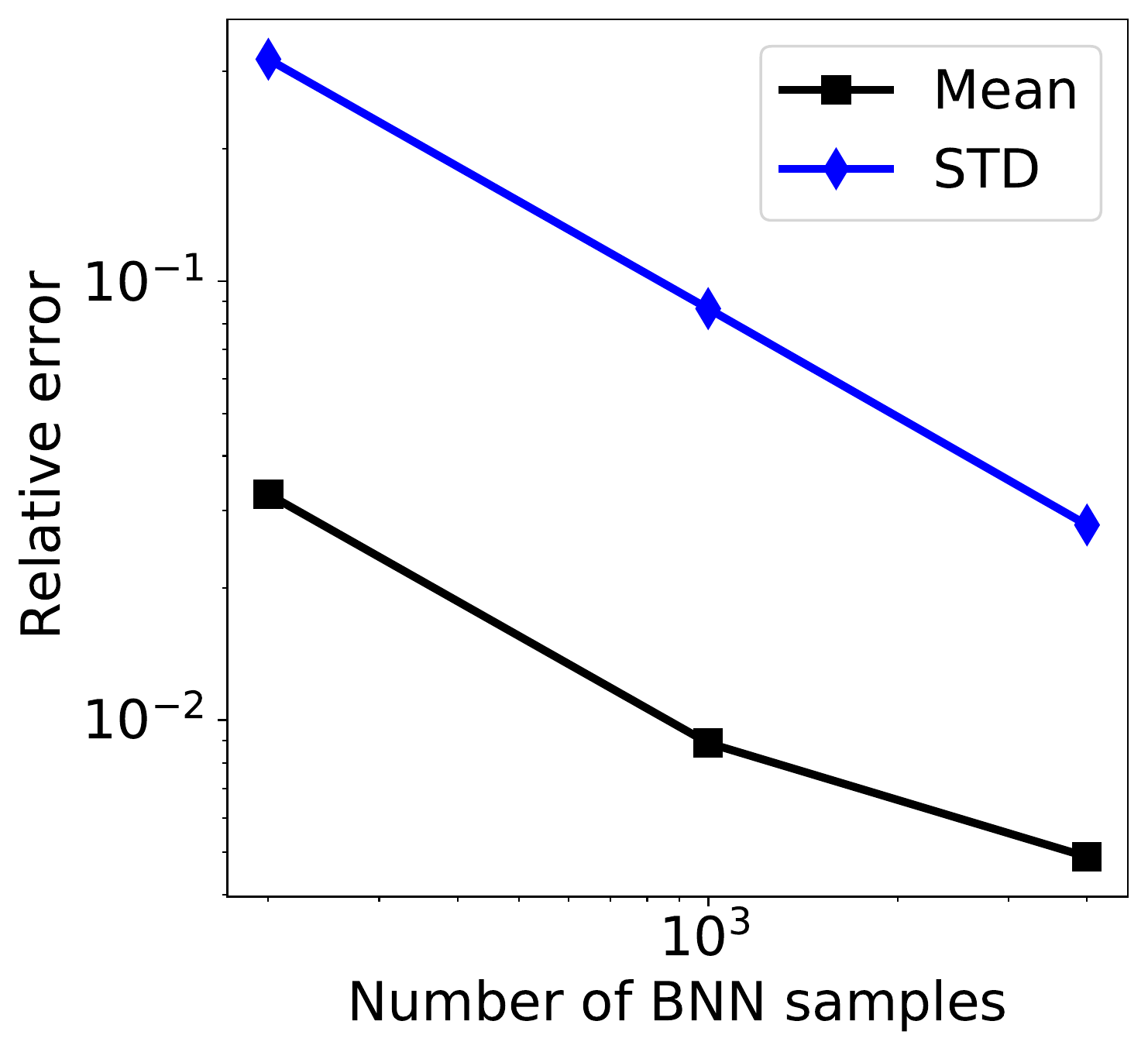}
        \caption{}
    \end{subfigure}
    \begin{subfigure}{0.40\textwidth}
        \centering
        \includegraphics[height=5cm]{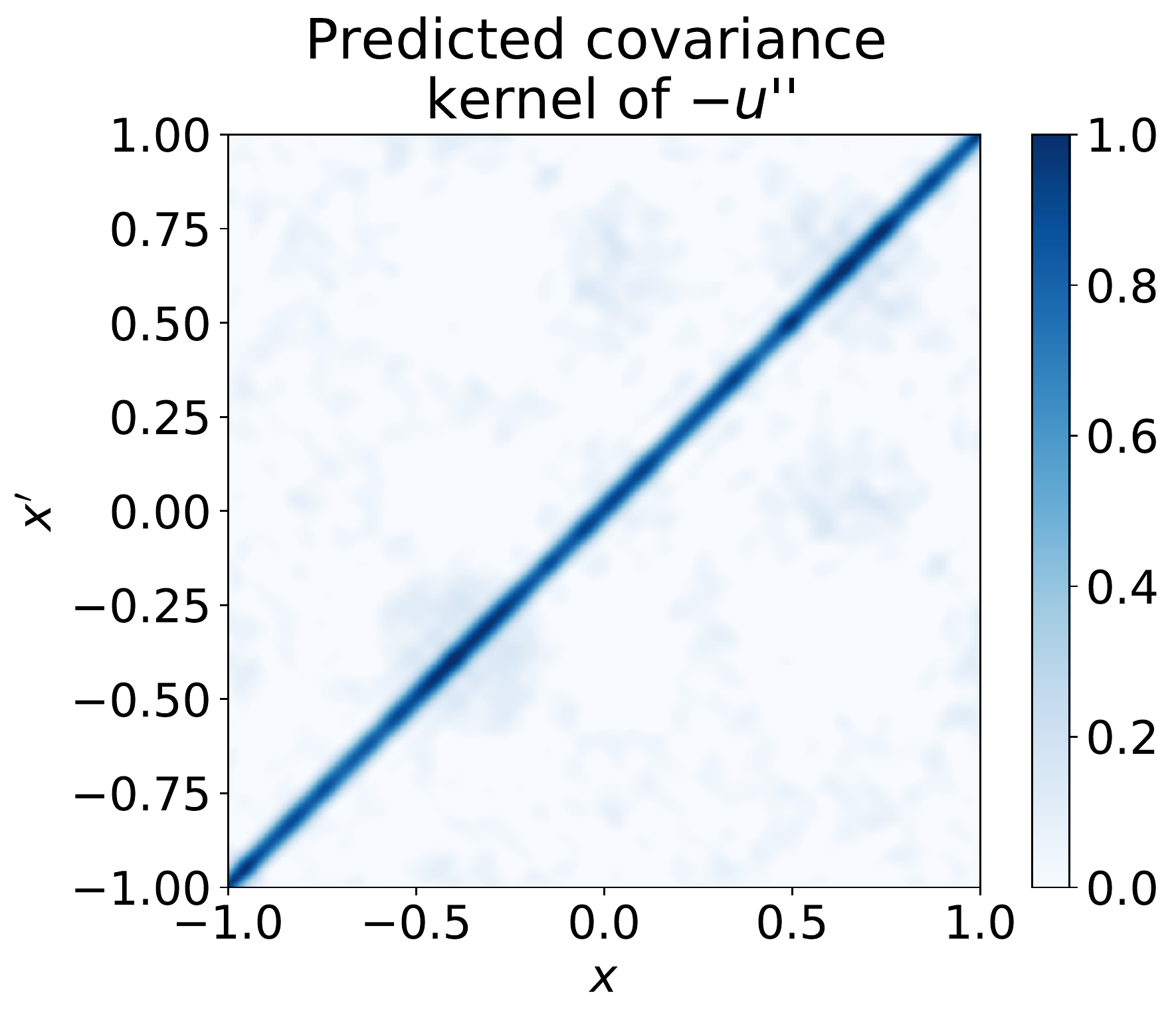}
        \caption{}
    \end{subfigure}
    \begin{subfigure}{0.40\textwidth}
        \centering
        \includegraphics[height=5cm]{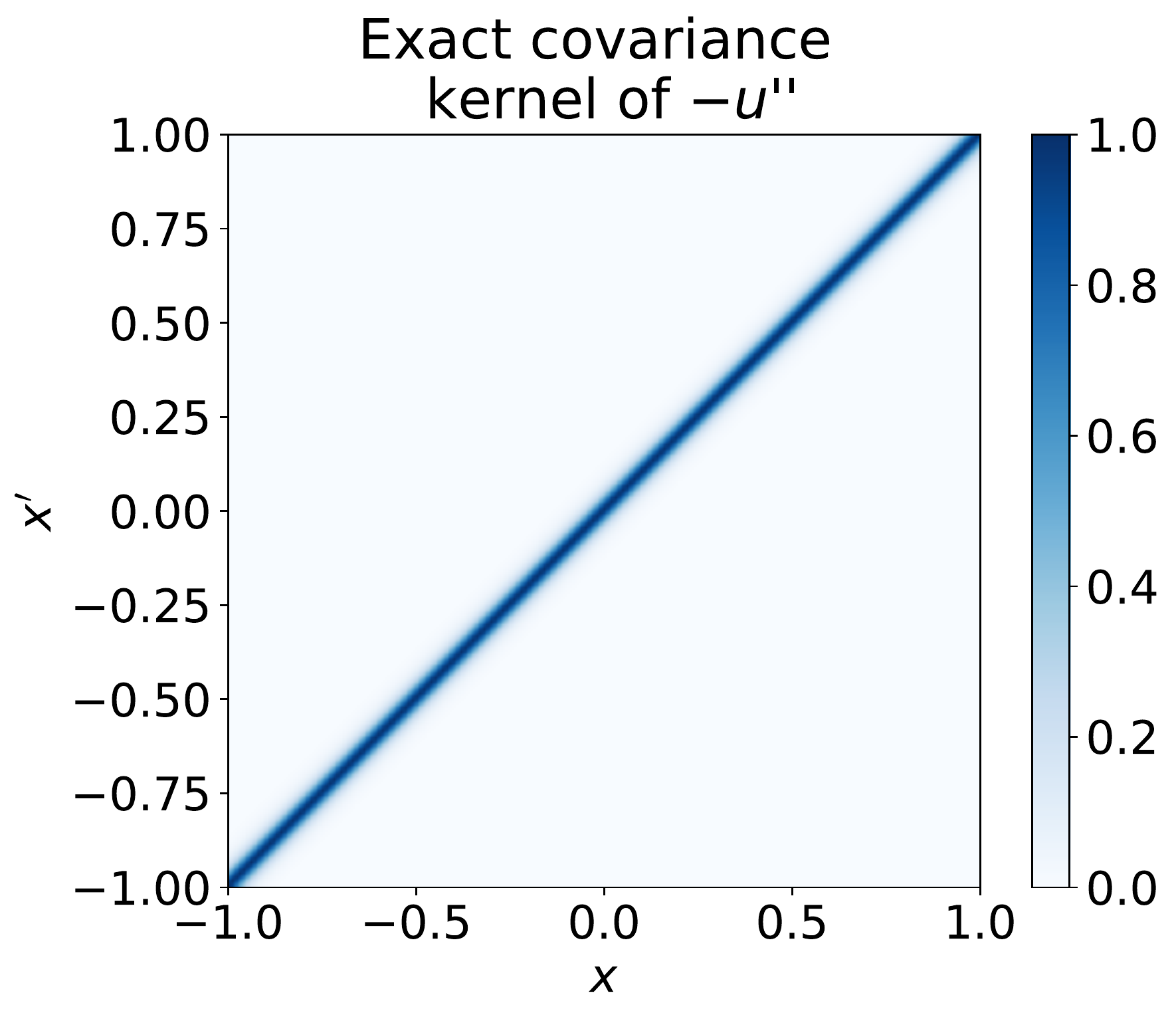}
        \caption{}
    \end{subfigure}

    \caption{Results of sampling the solution of the Poisson equation using BNN where $\sigma_f=1.0$ and $l_f=0.03$. The predicted (a) mean and (b) STD of the solution. Our proposed method worked well even for a high-dimensional problem. At the boundary, predicted STD is close to 0.01, which corresponds to the assumed noise STD. (c) Relative error of mean and STD vs number of BNN samples. (d) Predicted covariance kernel and (e) exact covariance kernel of $-u''$.}
   \label{fig:poisson_high}
    
\end{figure}

A small correlation length reduces the correlation between random processes at nearby points on the domain $D$, and thereby results in a high-frequency feature. To capture the feature, we assume an increase in the number of sensors on the domain, i.e., $N_f=101$. We also need an NN model that represents higher frequency; therefore, we use Fourier embeddings with $\sigma_1=1.0,\sigma_2=10.0$ and the rest of settings are the same. The mean, STD, and covariance of the negative second derivative of the solution give good approximations (Figure \ref{fig:poisson_high}). This result indicates that our proposed method works well even for a high-dimensional problem.

\begin{figure}[htbp]
    \centering
    \begin{subfigure}{0.99\textwidth}
        \centering
        \includegraphics[height=6cm]{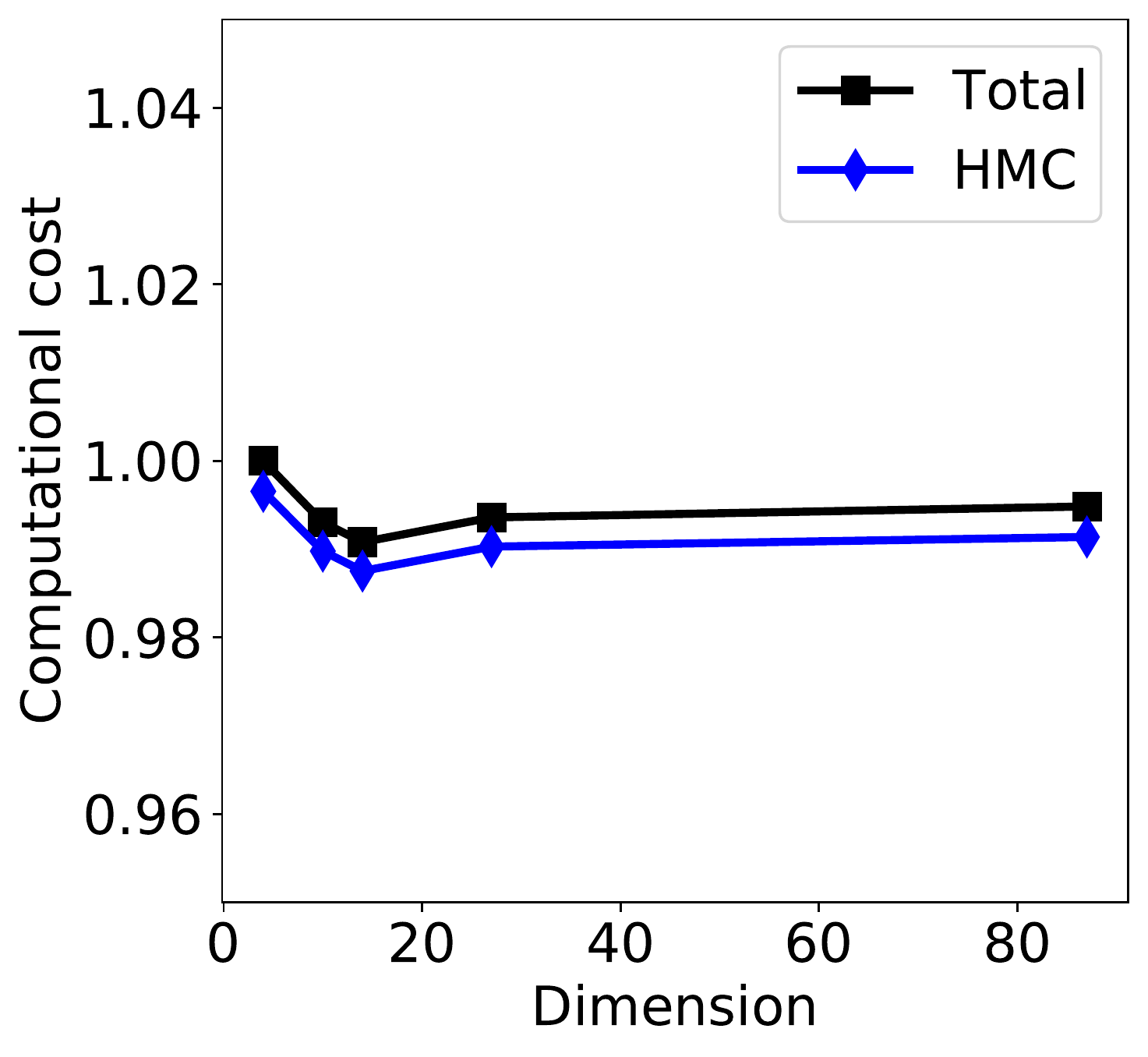}
    \end{subfigure}

   \caption{Normalized computational cost required to compute 4,000 BNN samples in each dimension, which does not increase with the growth of the dimension. $x$-axis: dimension that corresponds to each correlation length (Table \ref{tab:dimension}). $y$-axis: normalized computational time corresponding to the time required for $l_f = 1.0$, inducing the 4-dimensional problem.}
   \label{fig:poisson_time}
    
\end{figure}

We further investigate the computational cost required by our method for five correlation lengths of the source term. To be specific, we compare the time required to compute 4,000 BNN samples of the solution, using Algorithm \ref{alg:BNN_method} in each case. The computational cost does not increase with the growth of the dimension (Figure \ref{fig:poisson_time}). The increase in dimension changes the shape of the posterior distribution, which does not affect the cost of HMC, which dominates the overall cost. The result demonstrates that our method has the potential to solve the curse of dimensionality.

\subsection{The forward problem of 2D Allen-Cahn equation}

\begin{figure}[htbp]
    \centering
    \begin{subfigure}{0.40\textwidth}
        \centering
        \includegraphics[height=5.3cm]{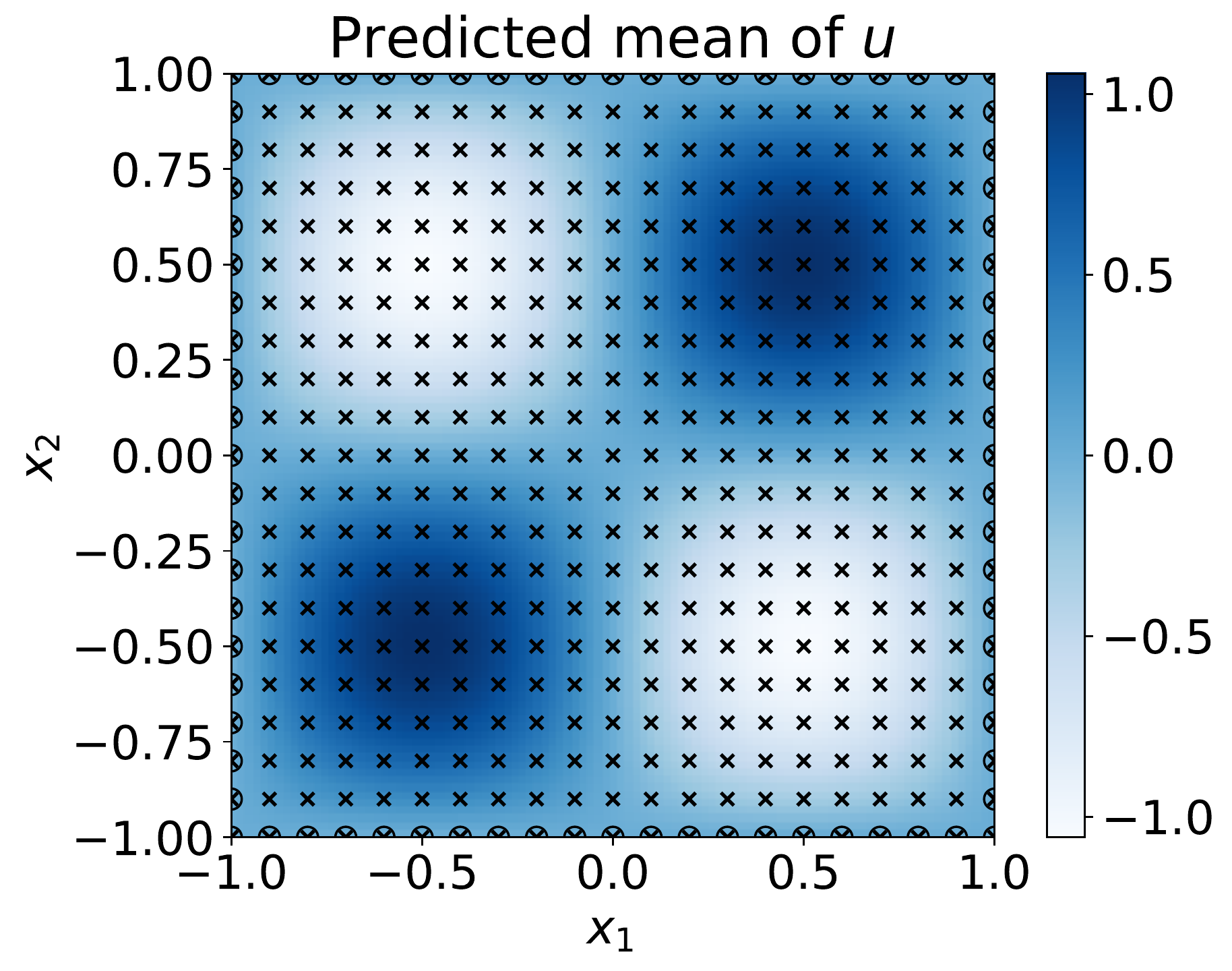}
        \caption{}
    \end{subfigure}
    \begin{subfigure}{0.40\textwidth}
        \centering
        \includegraphics[height=5.3cm]{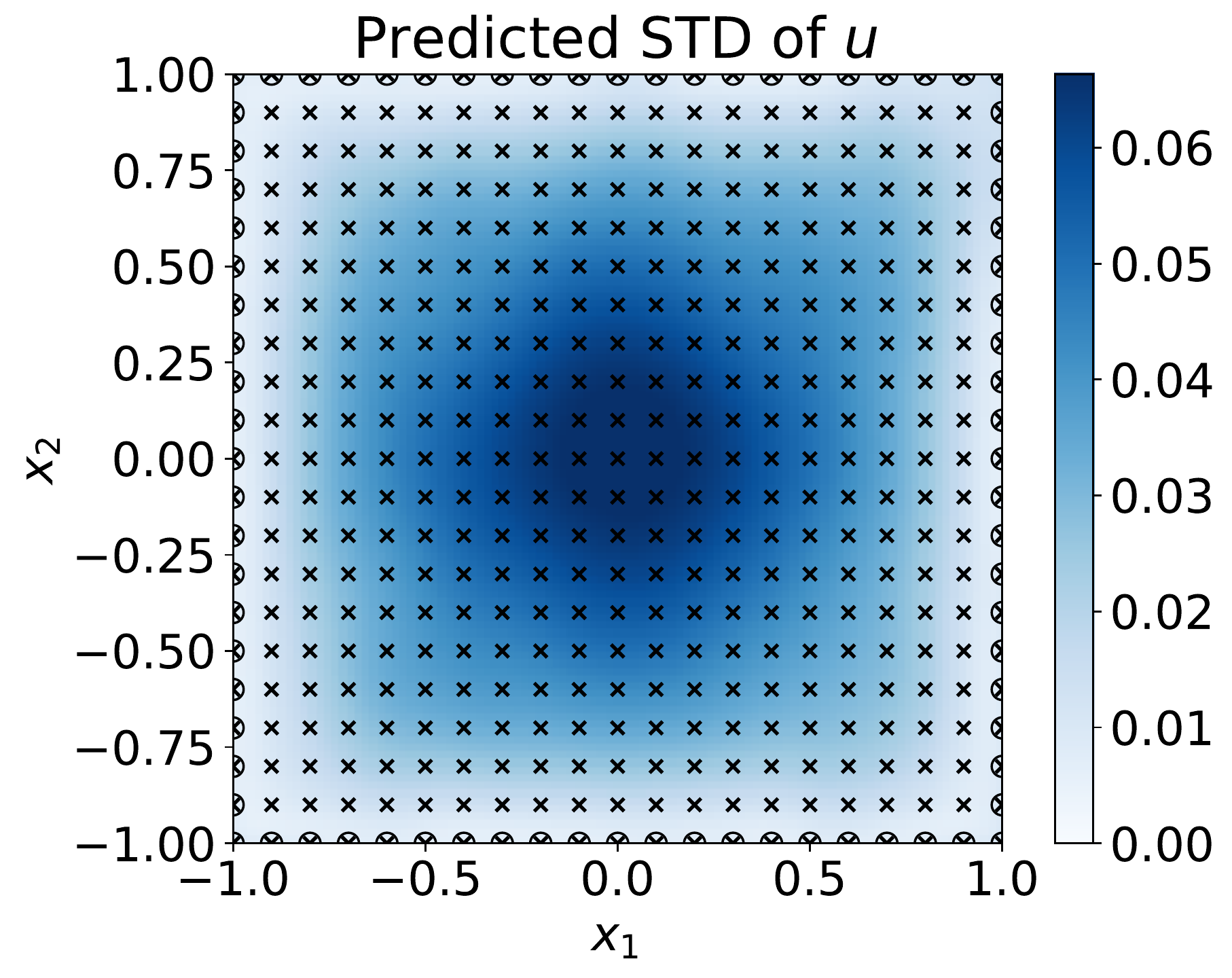}
        \caption{}
    \end{subfigure}
    \begin{subfigure}{0.40\textwidth}
        \centering
        \includegraphics[height=5.3cm]{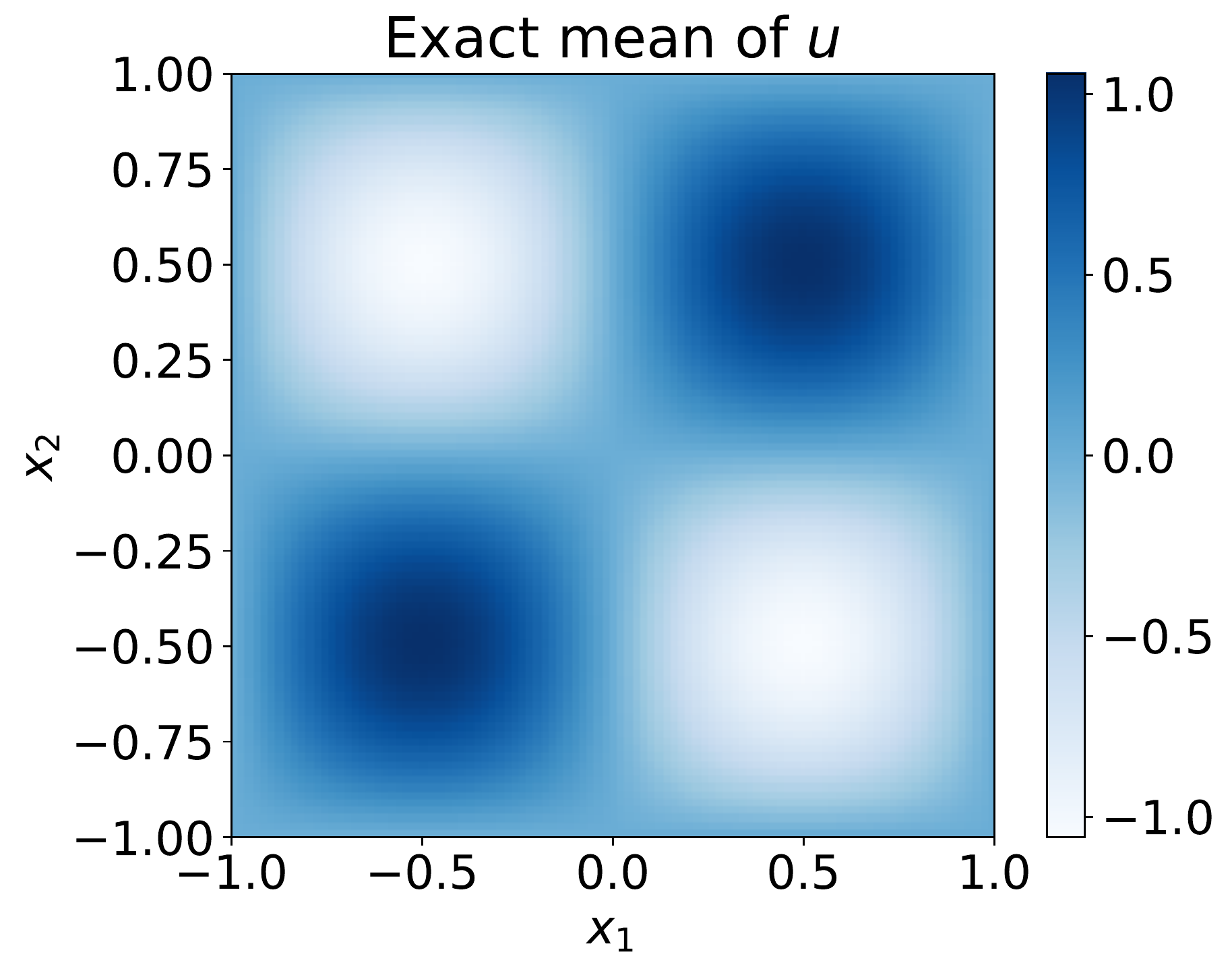}
        \caption{}
    \end{subfigure}
    \begin{subfigure}{0.40\textwidth}
        \centering
        \includegraphics[height=5.3cm]{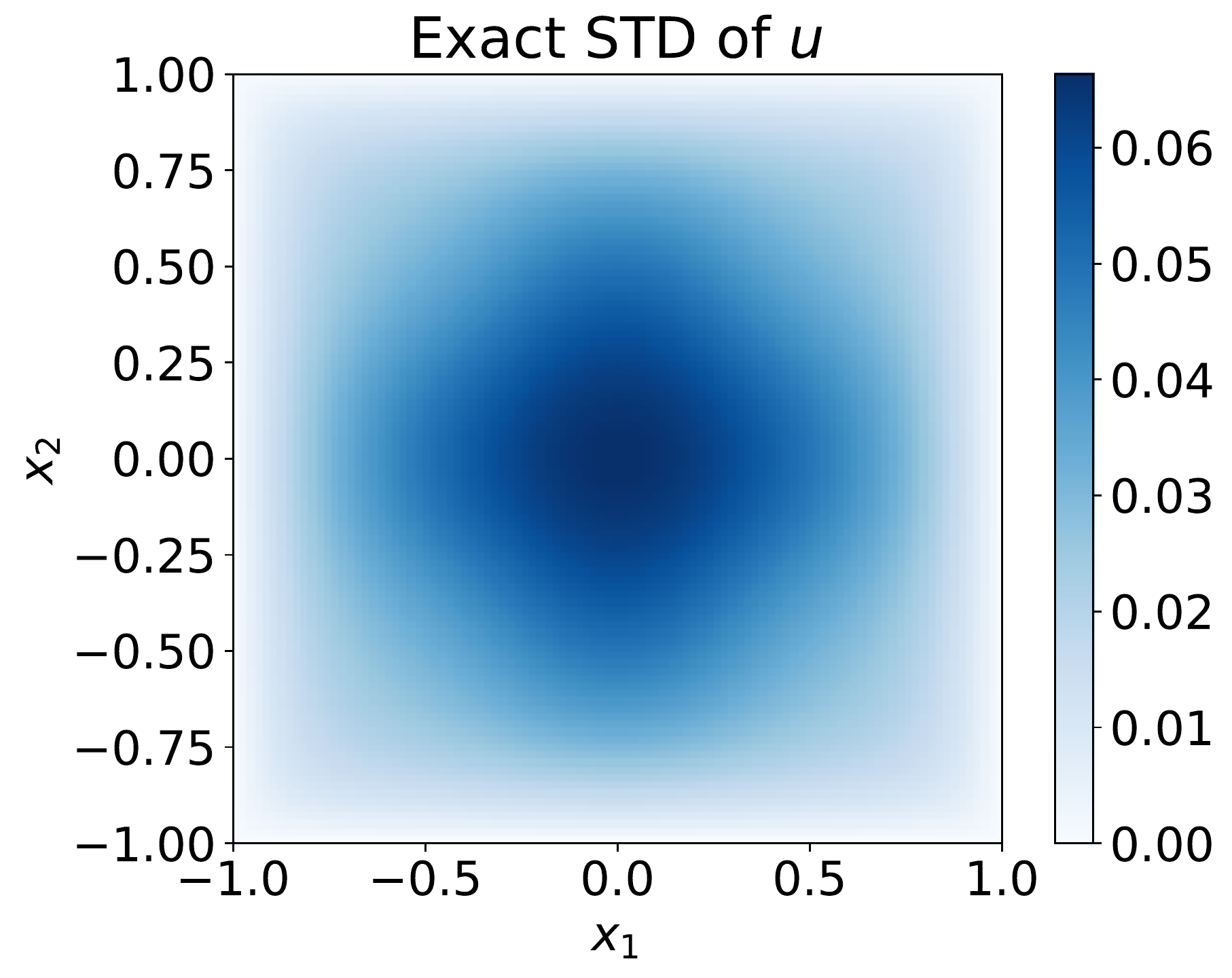}
        \caption{}
    \end{subfigure}
    \begin{subfigure}{0.32\textwidth}
        \centering
        \includegraphics[height=4.5cm]{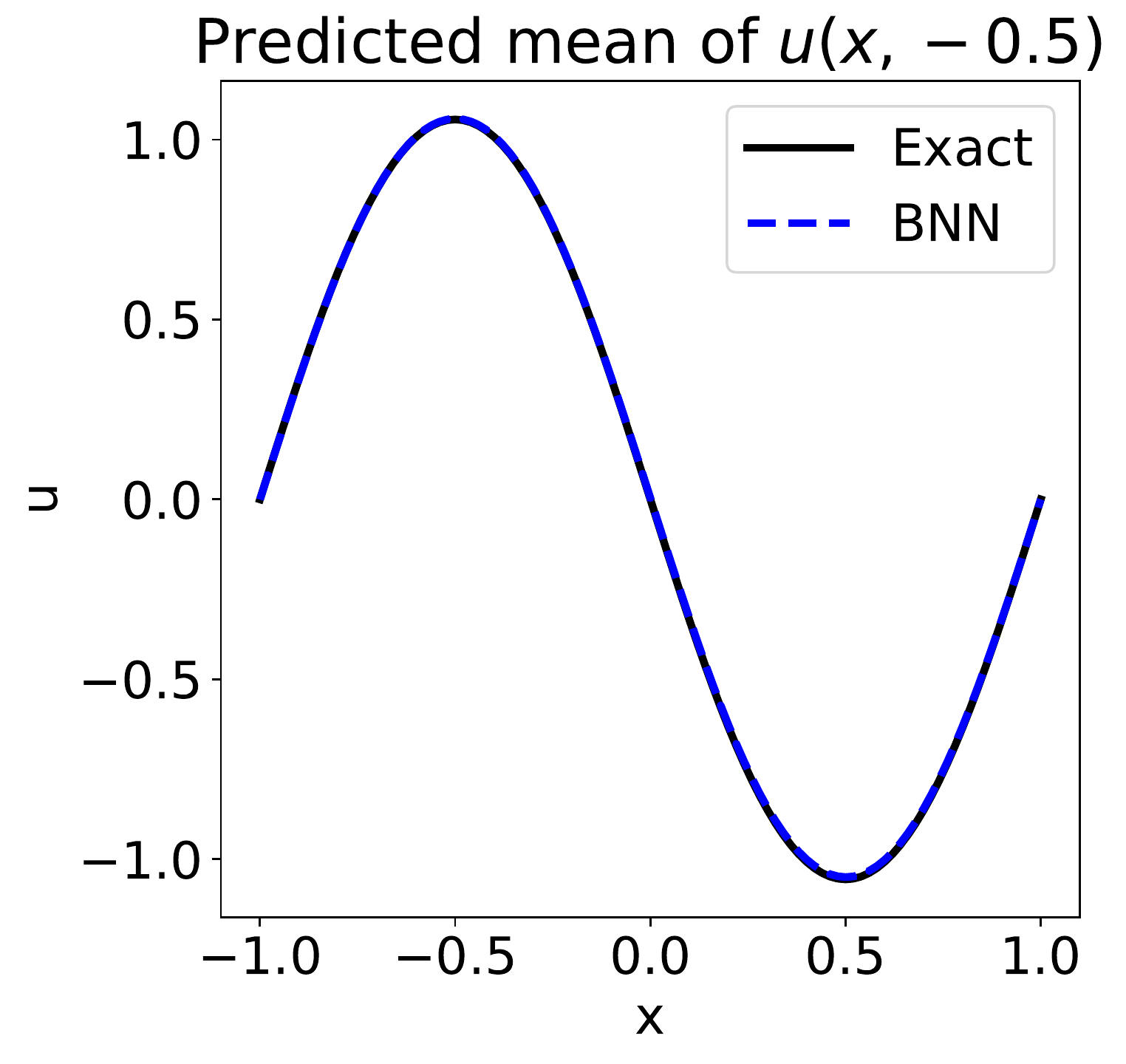}
        \caption{}
    \end{subfigure}
    \begin{subfigure}{0.32\textwidth}
        \centering
        \includegraphics[height=4.5cm]{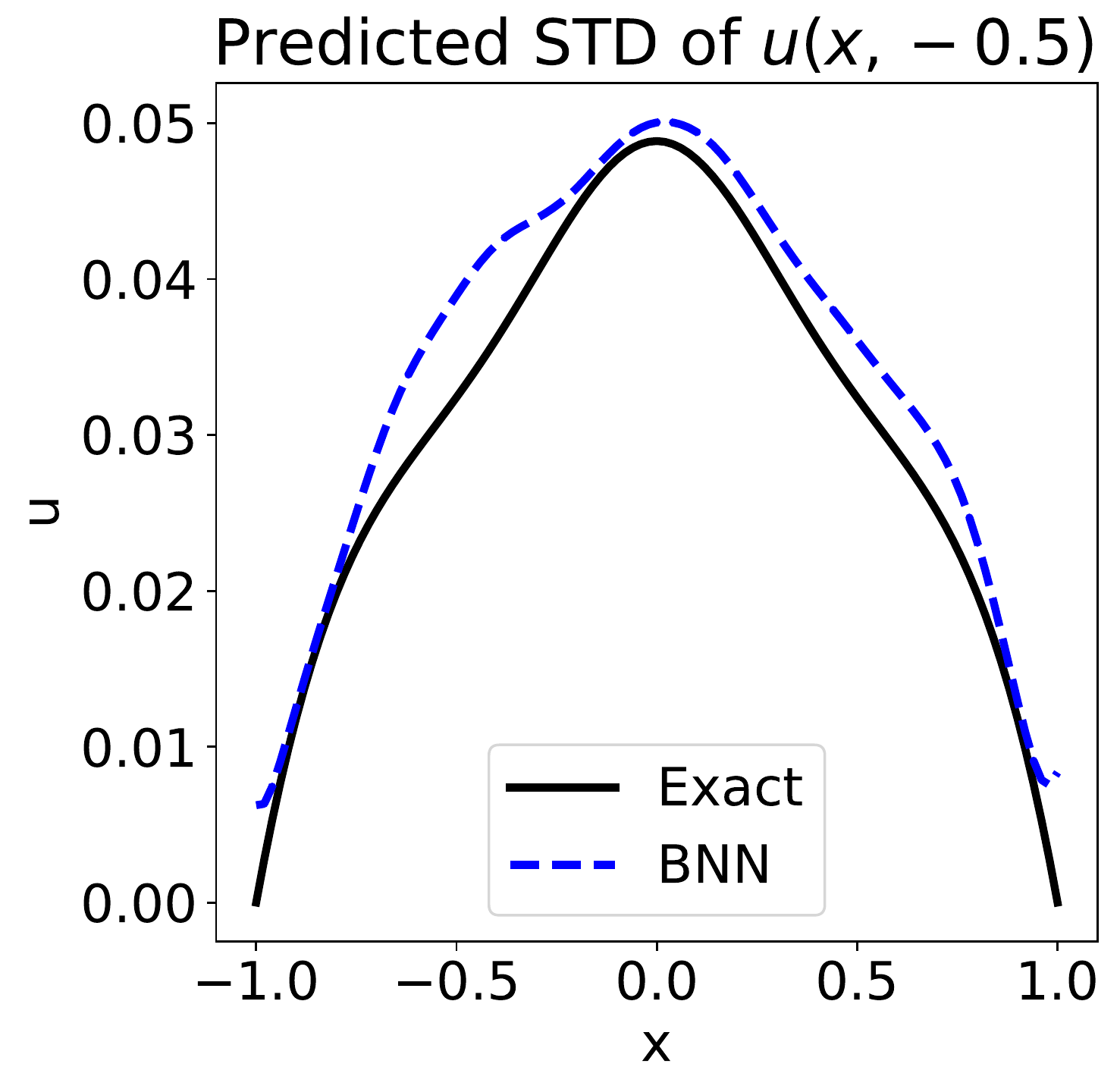}
        \caption{}
    \end{subfigure}
    \begin{subfigure}{0.32\textwidth}
        \centering
        \includegraphics[height=4.5cm]{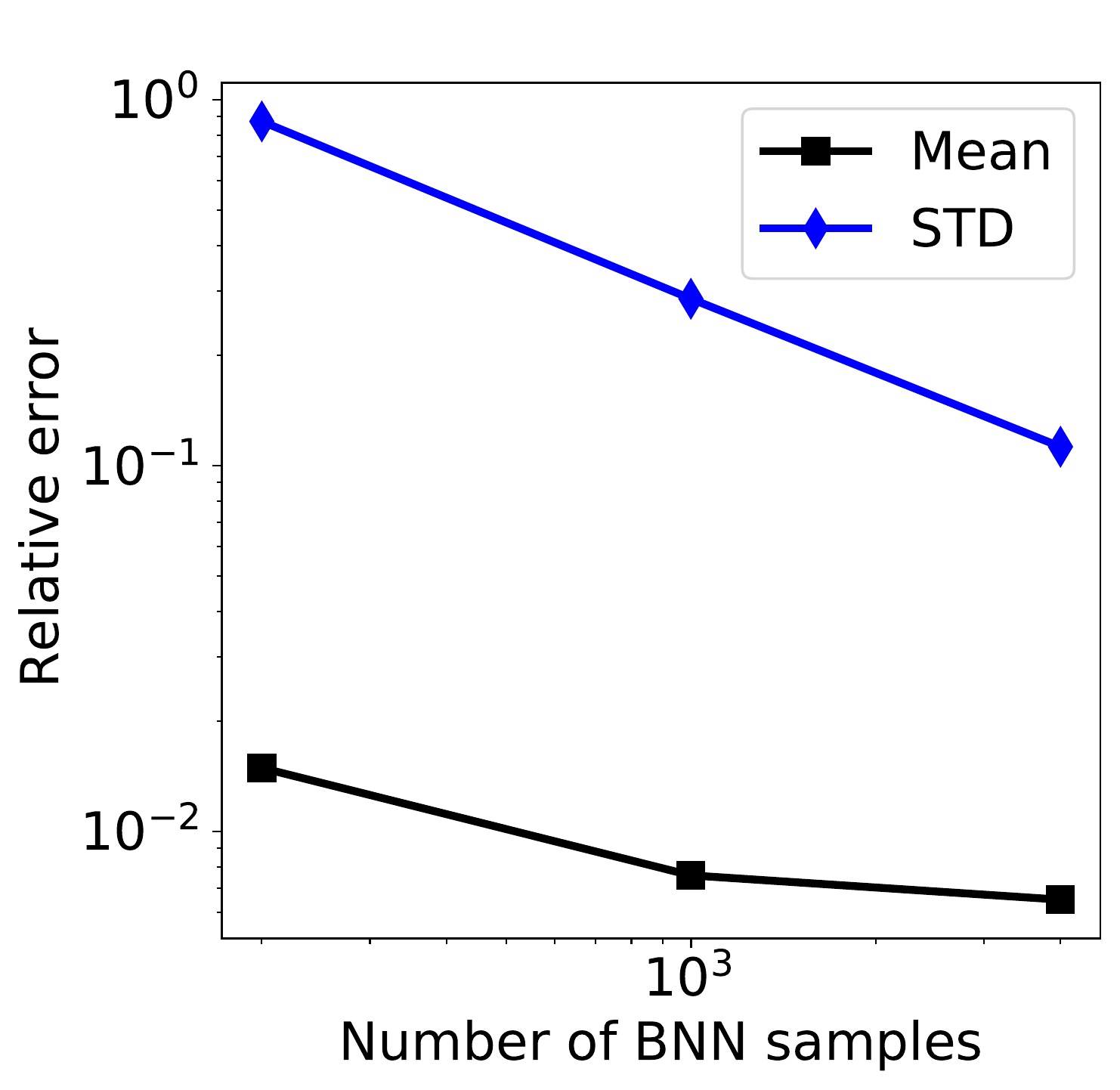}
        \caption{}
    \end{subfigure}

   \caption{The results of sampling the solution of the Allen-Cahn equation using BNN. Predicted (a) mean and (b) STD of the solution. Exact (c) mean and (d) STD of the solution. Predicted (e) mean and (f) STD of the solution for $x_2=-0.5$. At the boundary, STD is close to 0.01, which corresponds to the assumed noise STD. (g) Relative error of mean and STD vs number of BNN samples.}
   \label{fig:allencahn}
    
\end{figure}

We consider the 2D Allen-Cahn equation to apply our framework to the nonlinear example:
\begin{align*}
    -\Delta u + 3u(u^2-1) &= f\;\;\text{in $ D=[-1,1]^2$,} \\
    u &= 0\;\;\text{on $\partial D$.}
\end{align*}
It describes the process of phase separation in multi-component alloy systems, including order-disorder transitions. The distribution of $f$ is set to
\begin{equation*}
f(\mathbf{x}) \sim \mathcal{GP}(20\sin \pi x_1 \sin \pi x_2, \sigma_f^2 \exp \Big(-\frac{|\mathbf{x}-\mathbf{x}'|^2}{2 l_f^2}\Big))\;\;\text{for $\mathbf{x}=(x_1,x_2) \in D$},
\end{equation*}
where $\sigma_f=1.0$ and $l_f=0.1$. $N_f=441$ sensors of $f$ are placed in the domain (Figure \ref{fig:allencahn}, `x' marks). In addition, $N_u=80$ sensors of $u$ are placed equidistantly around the borders of the domain (Figure \ref{fig:allencahn}, `o' marks). We assume that each sensor of $u$ is subject to measurement noise that has a normal distribution $\mathcal{N}(0,0.01^2)$.

We use an NN model that is a multiscale FFN with two Fourier feature embeddings where $B^{(i)} \in \mathbb{R}^{50 \times 2}, \sigma_1=1.0$, and $\sigma_2=5.0$. The embedded inputs pass through an FCN that has a sine activation function, and that consists of one hidden layer of with 200 units. In the HMC setup, the number of leapfrog steps is $M=\text{2,000}$, and the time step is $\delta=5 \times 10^{-6}$. The results provide good predictions of mean and STD (Figure \ref{fig:allencahn}). As in Section \ref{sec:1dpoisson}, the noise assumed at the boundary appears as an error of the STD at the boundary. We conclude that BNN provides an integrated method to solve both linear and nonlinear SPDEs.

\subsection{The inverse problem of 1D elliptic equation}

\begin{figure}[htbp]
    \centering
    \begin{subfigure}{0.32\textwidth}
        \centering
        \includegraphics[height=4.5cm]{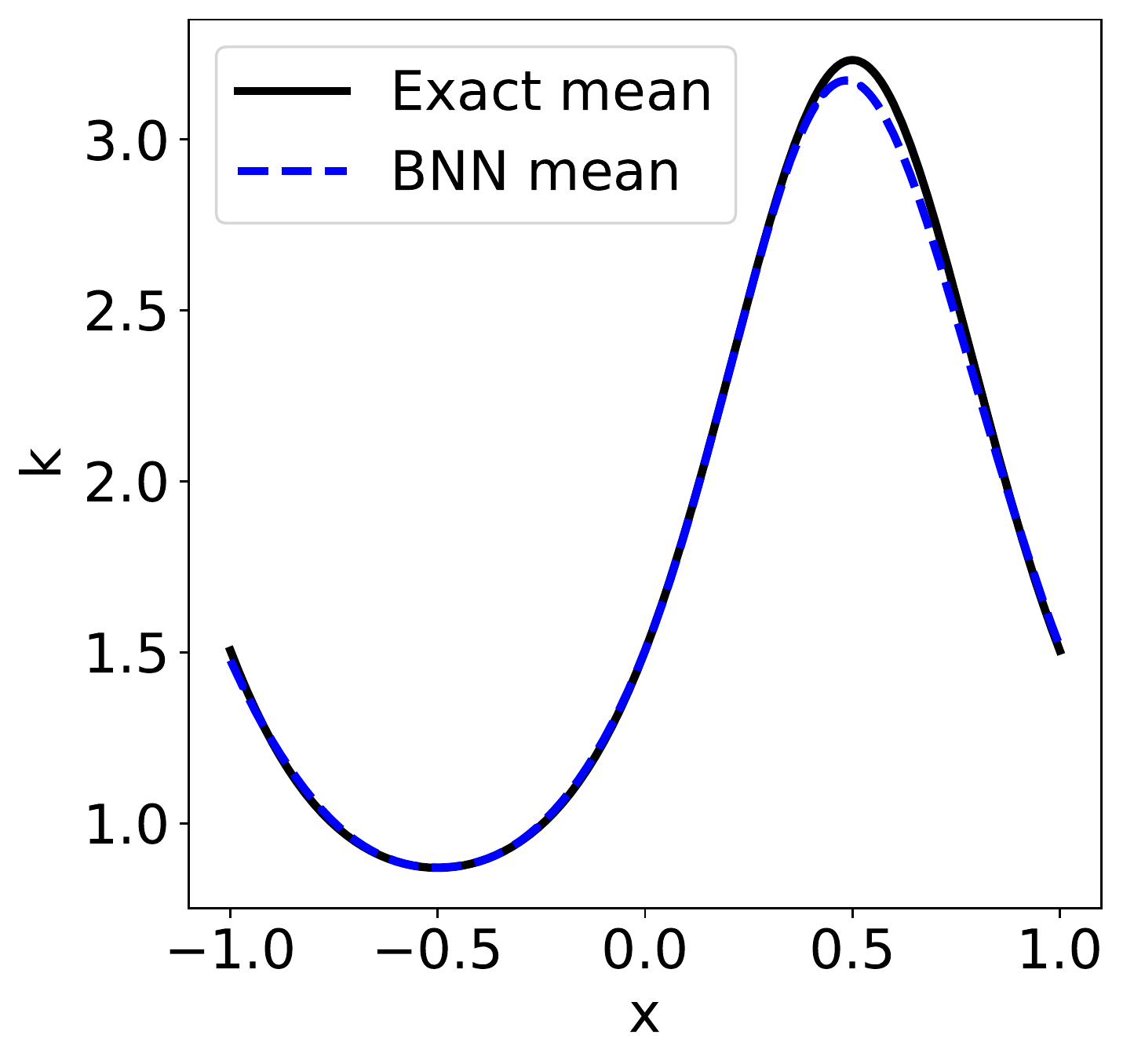}
        \caption{}
    \end{subfigure}
    \begin{subfigure}{0.32\textwidth}
        \centering
        \includegraphics[height=4.5cm]{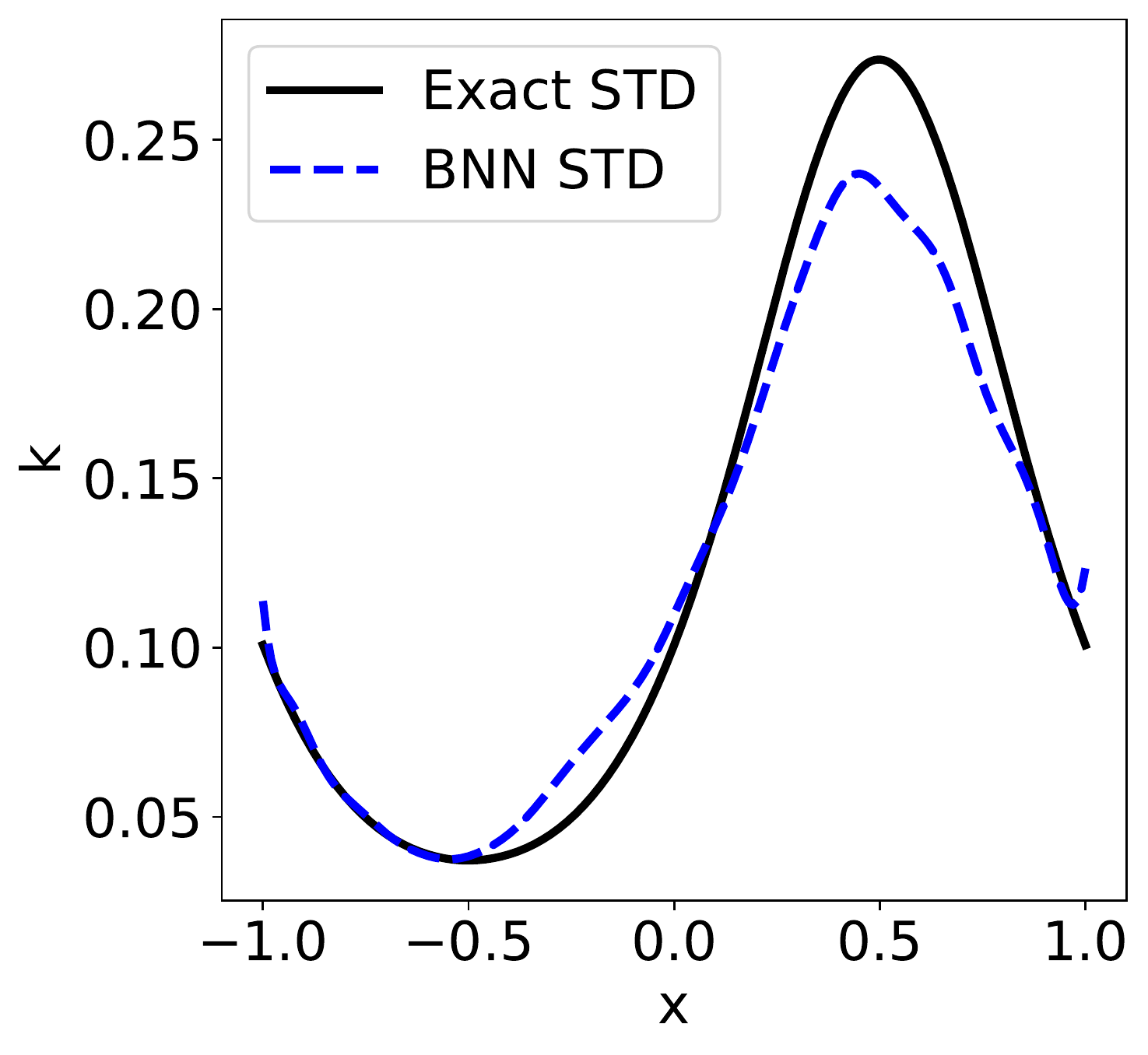}
        \caption{}
    \end{subfigure}
    \begin{subfigure}{0.32\textwidth}
        \centering
        \includegraphics[height=4.5cm]{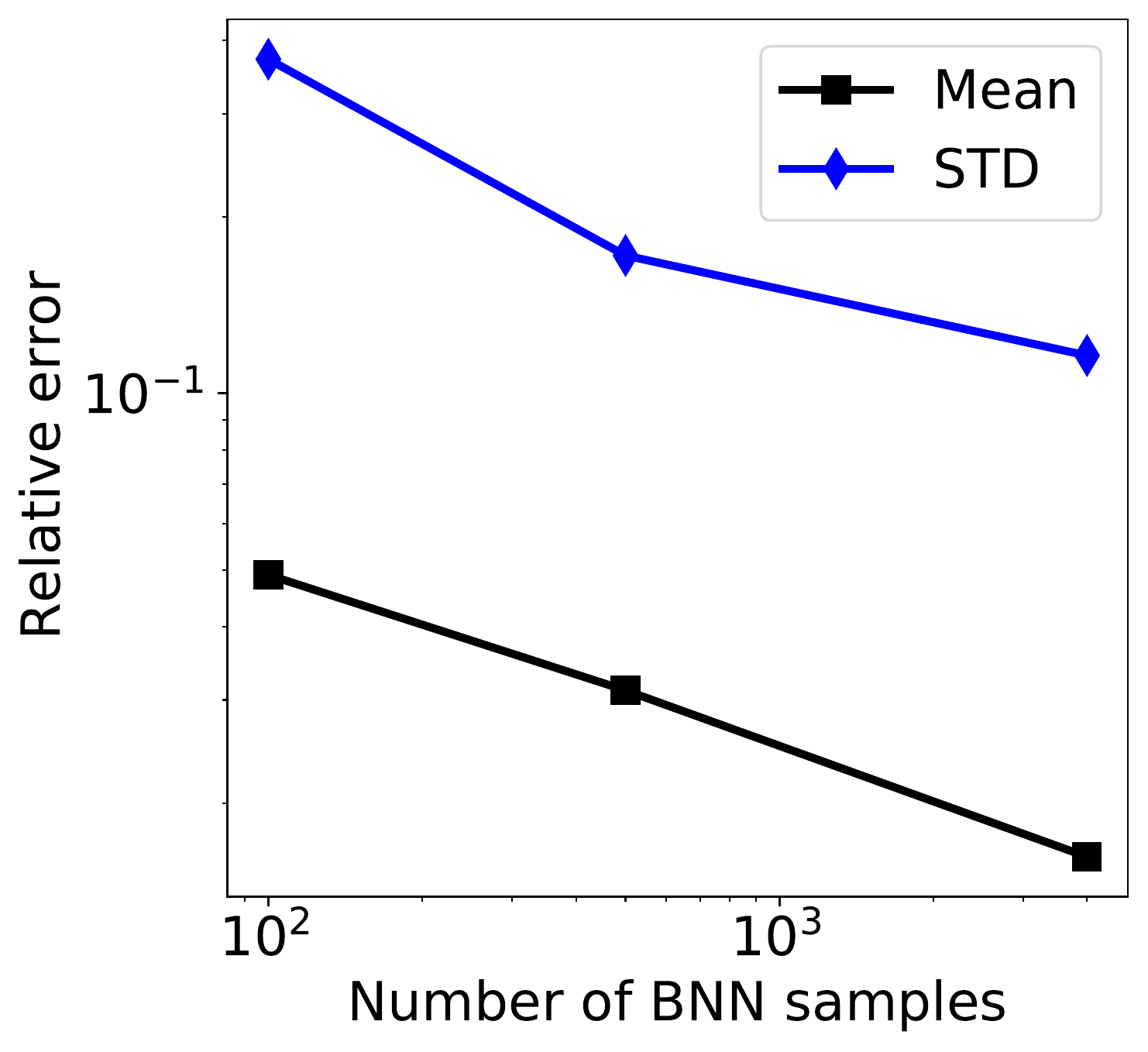}
        \caption{}
    \end{subfigure}
    \begin{subfigure}{0.40\textwidth}
        \centering
        \includegraphics[height=5cm]{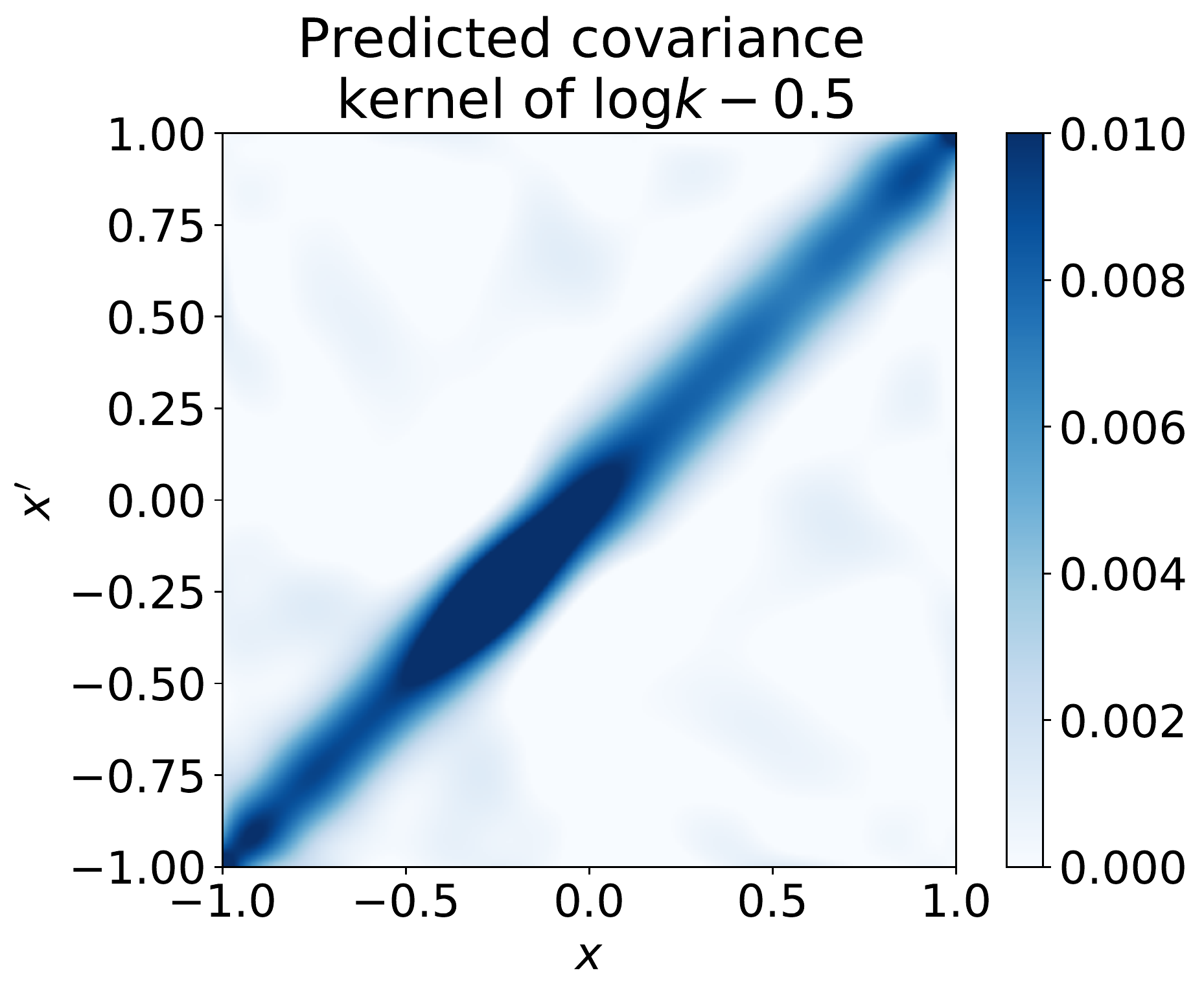}
        \caption{}
    \end{subfigure}
    \begin{subfigure}{0.40\textwidth}
        \centering
        \includegraphics[height=5cm]{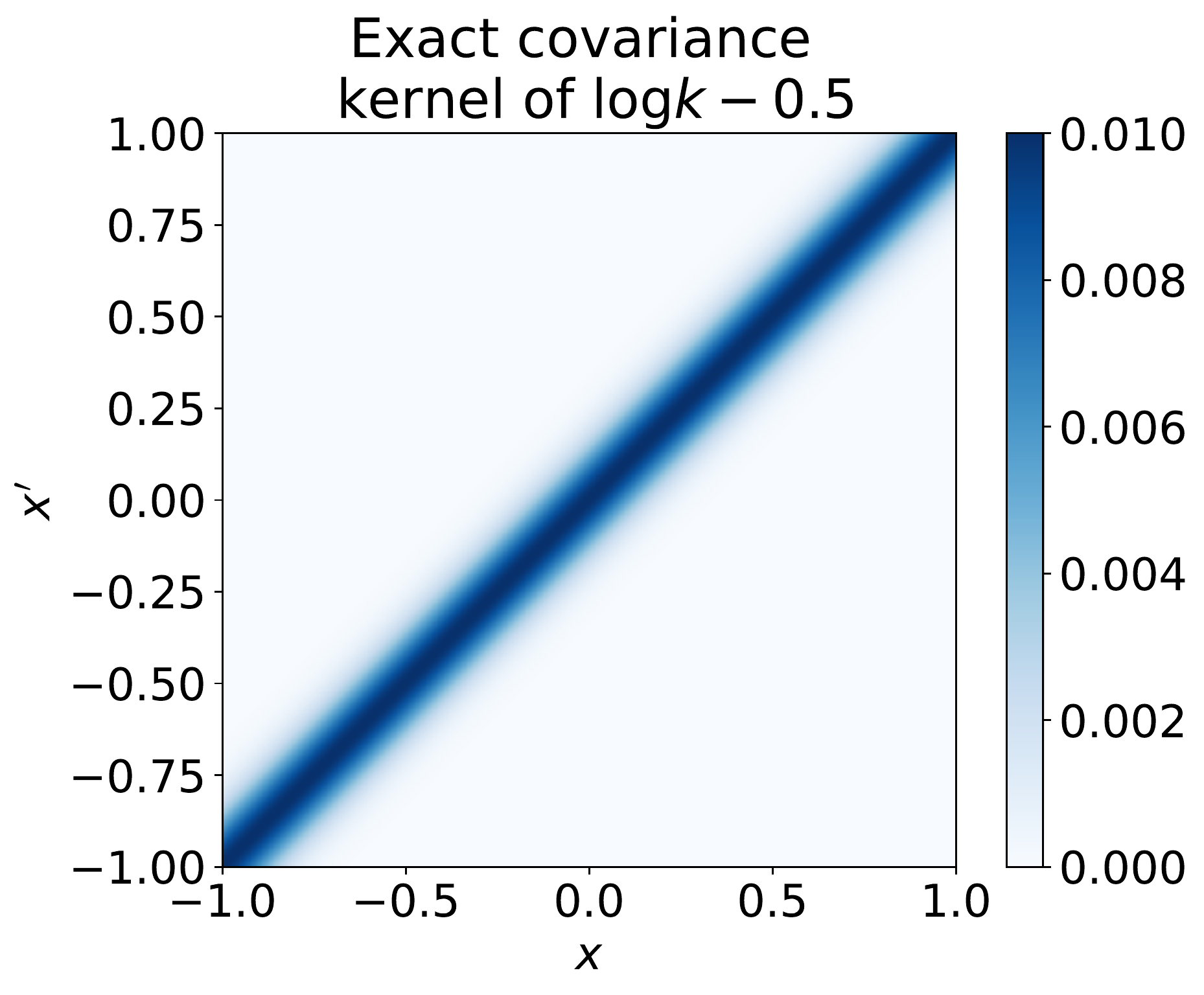}
        \caption{}
    \end{subfigure}

    \caption{Results of sampling the parameter $k$ of the elliptic equation using BNN where $\sigma_k=0.1,l_k=0.1,\sigma_f=0.3$, and $l_f=0.1$. Predicted (a) mean and (b) STD of the samples. (c) Relative error of mean and STD vs number of BNN samples. (d) Predicted covariance kernel and (e) exact covariance kernel of $\log k - 0.5$.}
   \label{fig:elliptic_low}
    
\end{figure}

In this section, we infer the unknown parameter contained in the differential operator of the elliptic equation on a 1D domain:
\begin{align*}
    -\frac{d}{dx}(k(x,\omega)\frac{d}{dx}u(x,\omega)) &= f(x,\omega)\;\;\text{for $x \in D=[-1,1]$}, \\
    u(-1),u(1) &= 0, \\
\end{align*}
where the distributions of $k$ and $f$ are independent and defined as
\begin{align*}
    \log k(x) - 0.5 &\sim \mathcal{GP}(\sin \pi x, \sigma_k^2 \exp \Big(-\frac{|x-x'|^2}{2 l_k^2}\Big))\;\;\text{for $x \in D=[-1,1]$}, \\
    f(x) &\sim \mathcal{GP}(3, \sigma_f^2 \exp \Big(-\frac{|x-x'|^2}{2 l_f^2}\Big))\;\;\text{for $x \in D=[-1,1]$},
\end{align*}
with $\sigma_k=0.1,l_k=0.1,\sigma_f=0.3$, and $l_f=0.1$. The diffusion coefficient $k$ is assumed to be unknown and inferred from the measurements of $u$ and $f$ on the spatial sensors. There are $N_u=41$ and $N_f=41$ equidistant sensors on the domain for both $u$ and $f$, respectively. We use a multiscale FFN with two Fourier feature embeddings where $B^{(i)} \in \mathbb{R}^{10 \times 1},\sigma_1=1.0$, and $\sigma_2=5.0$. The embedded inputs pass through an FCN that uses a sine activation function, and consists of one hidden layer of with 200 units. The number of leapfrog step is $M=300$, and the step size is $\delta=3 \times 10^{-5}$. The predicted values of the mean, STD, and covariance agree well with reference values (Figure \ref{fig:elliptic_low}).

\begin{figure}[htbp]
    \centering
    \begin{subfigure}{0.32\textwidth}
        \centering
        \includegraphics[height=4.5cm]{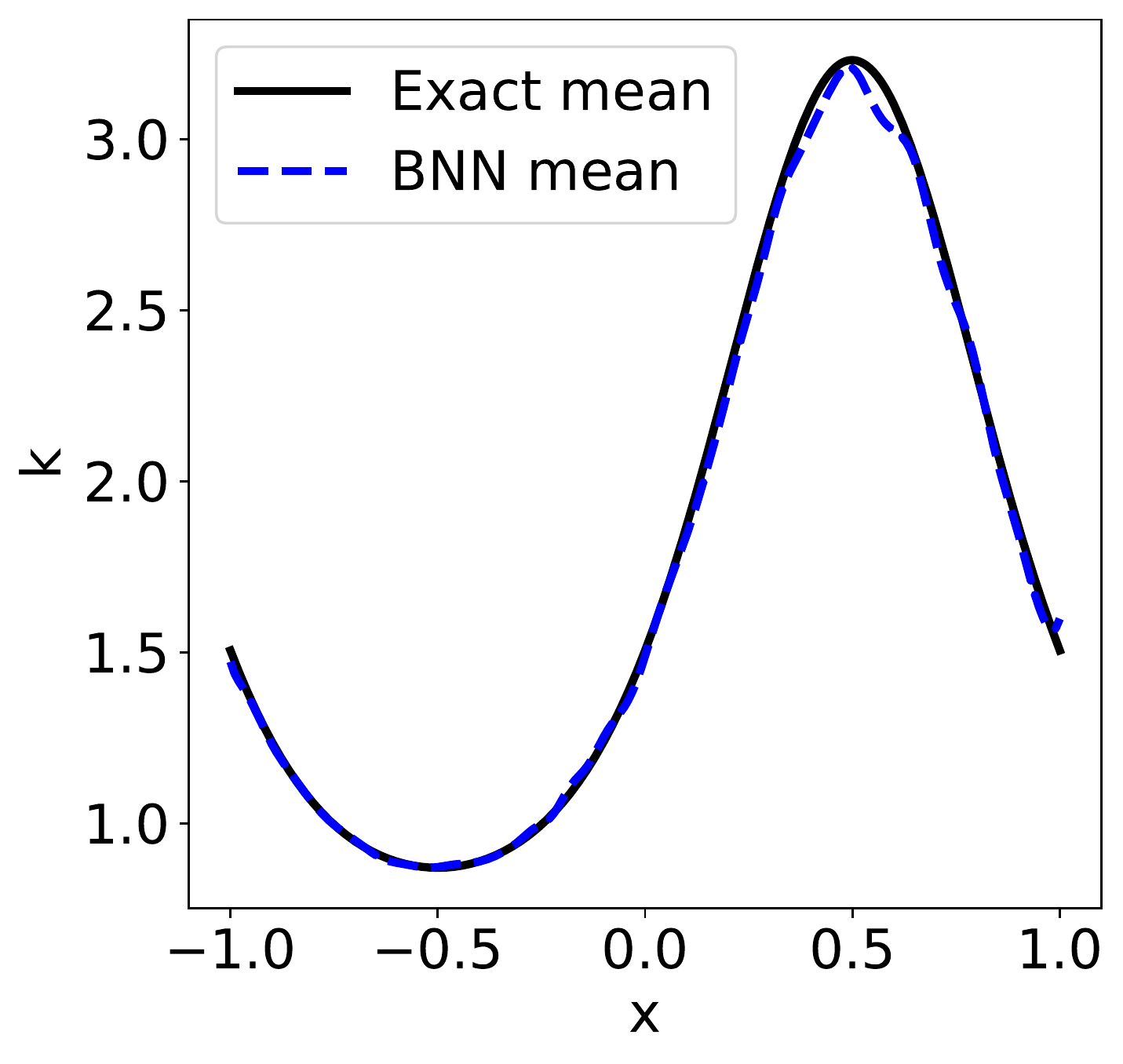}
        \caption{}
    \end{subfigure}
    \begin{subfigure}{0.32\textwidth}
        \centering
        \includegraphics[height=4.5cm]{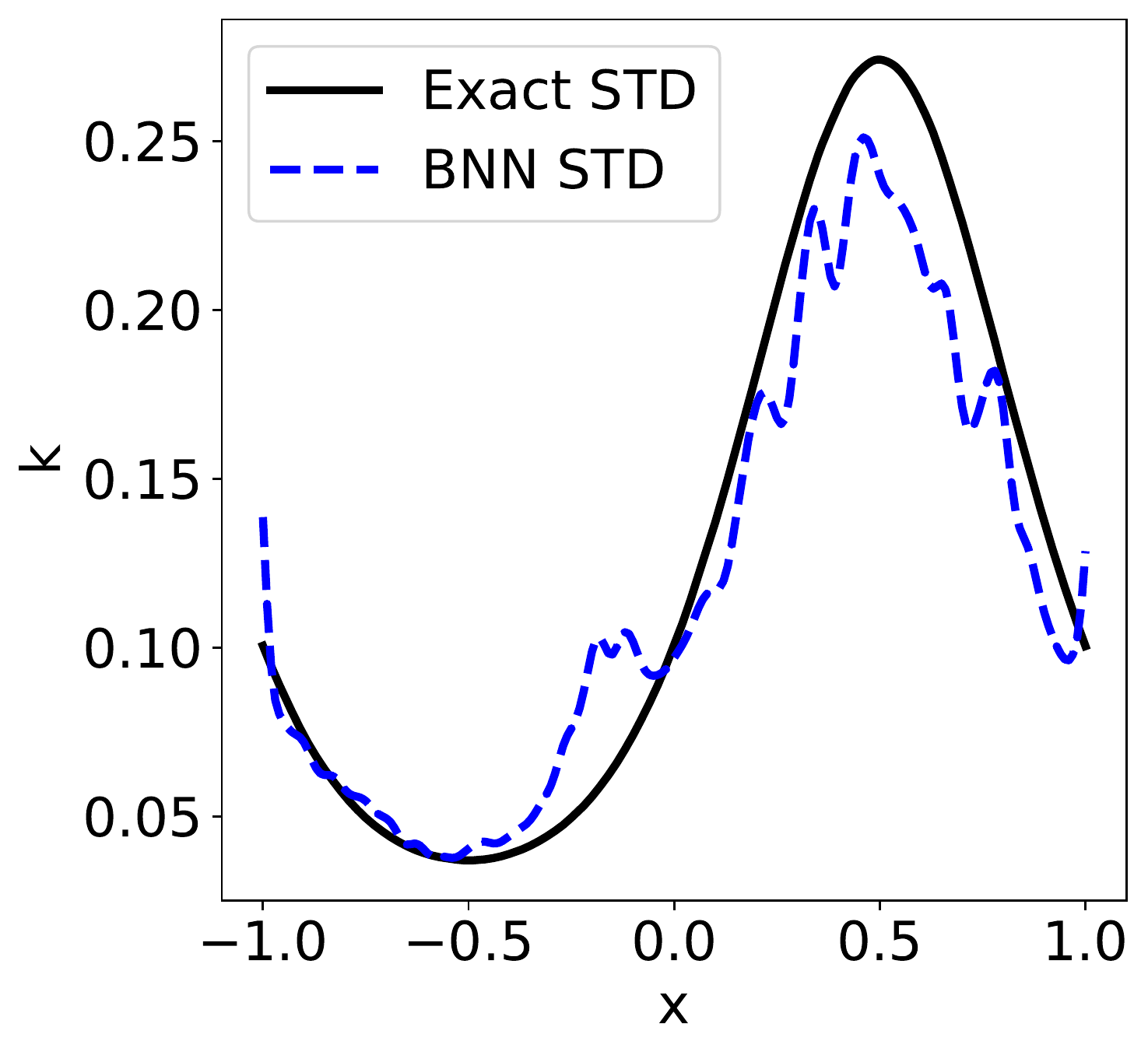}
        \caption{}
    \end{subfigure}
    \begin{subfigure}{0.32\textwidth}
        \centering
        \includegraphics[height=4.5cm]{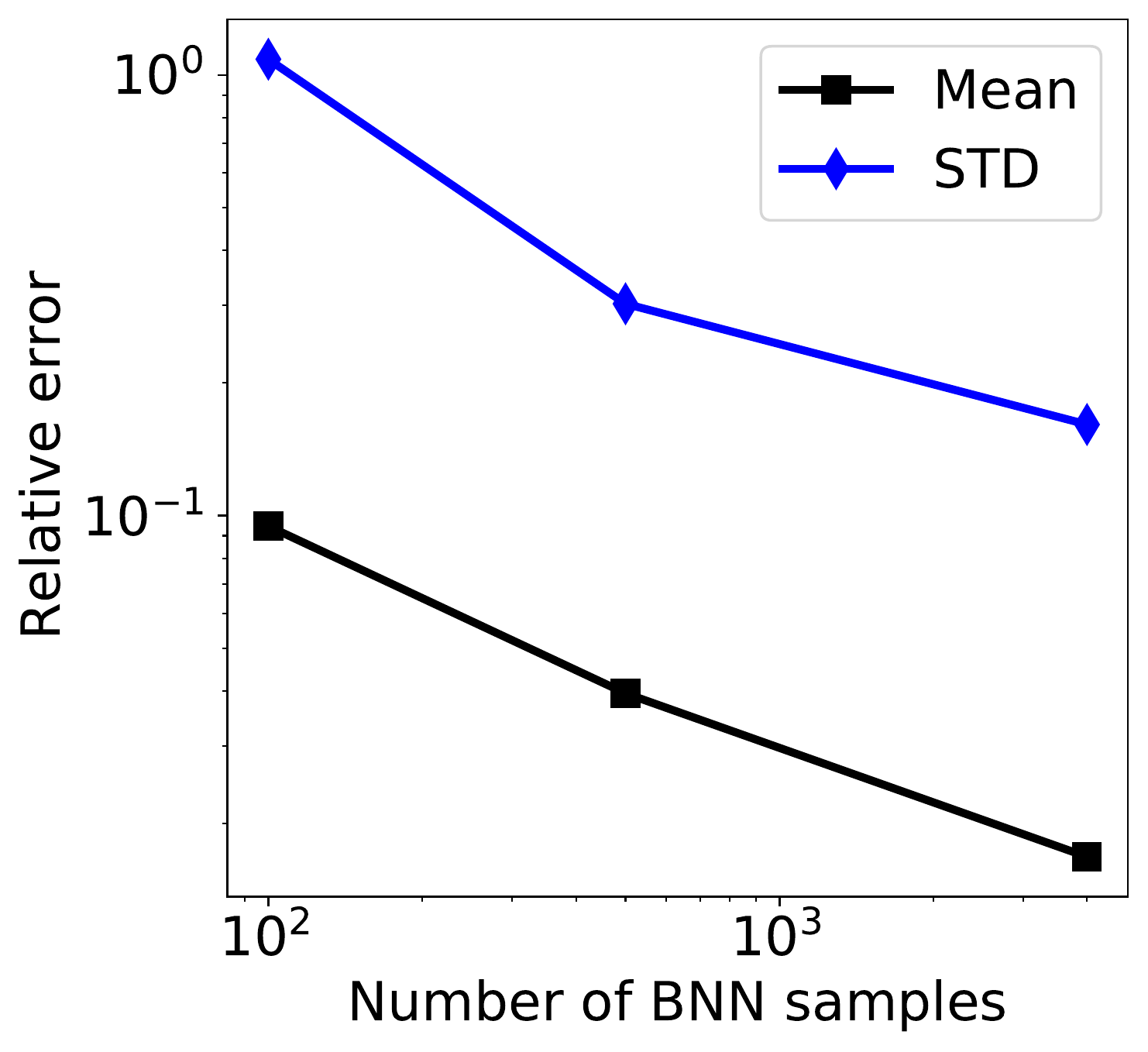}
        \caption{}
    \end{subfigure}
    \begin{subfigure}{0.40\textwidth}
        \centering
        \includegraphics[height=5cm]{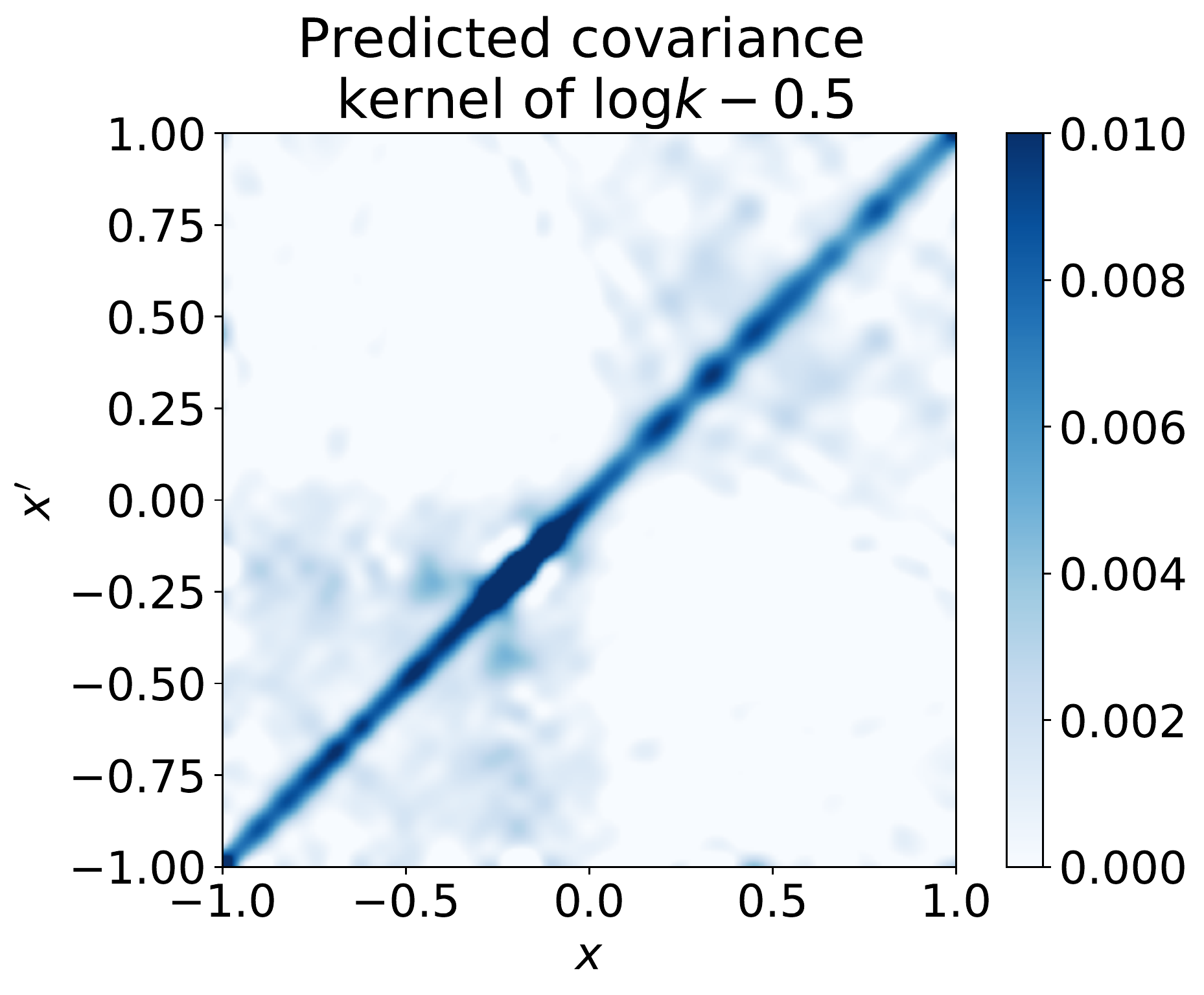}
        \caption{}
    \end{subfigure}
    \begin{subfigure}{0.40\textwidth}
        \centering
        \includegraphics[height=5cm]{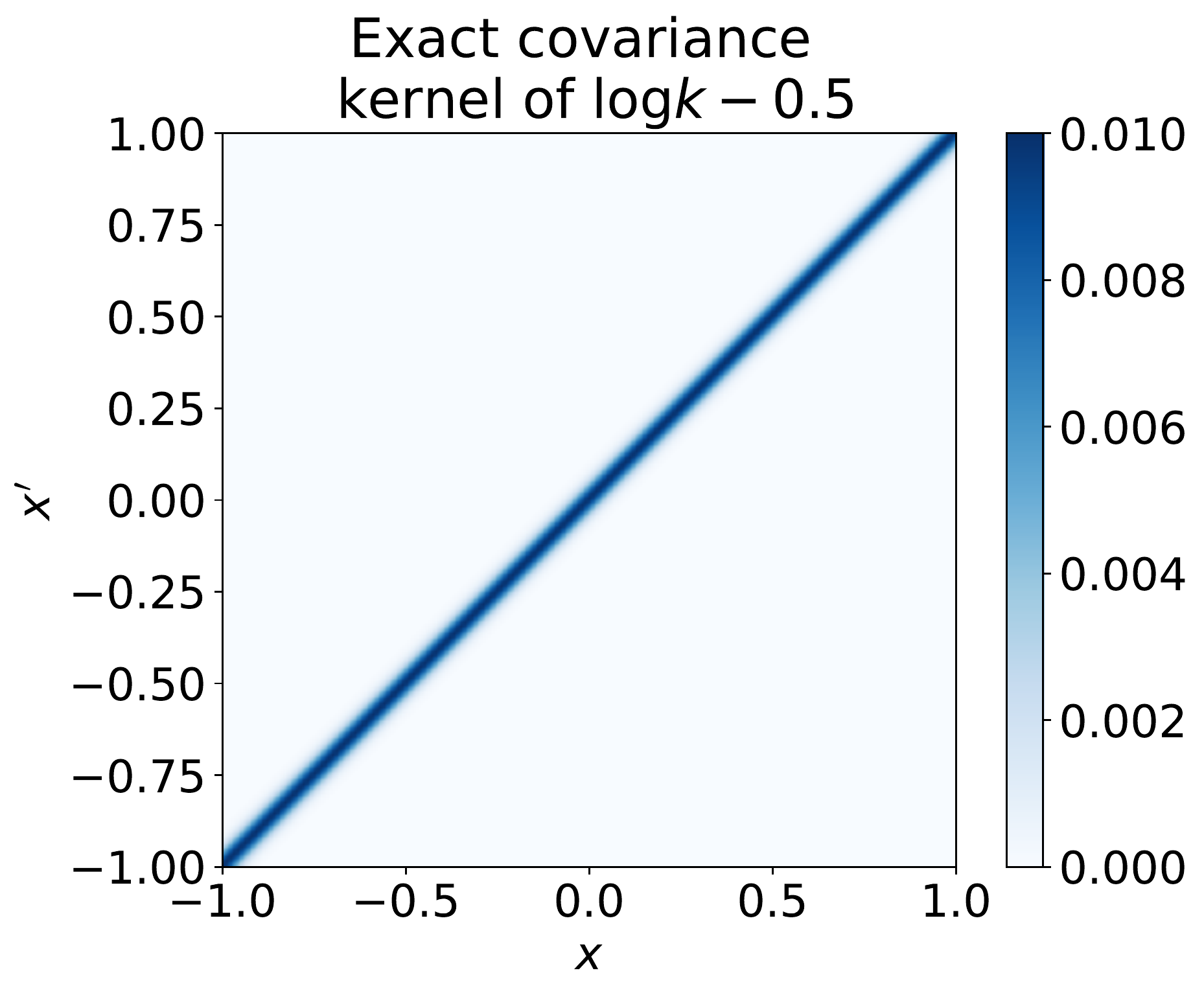}
        \caption{}
    \end{subfigure}

    \caption{Results of sampling the parameter $k$ of the elliptic equation by using a BNN where $\sigma_k=0.1,l_k=0.03,\sigma_f=0.3$, and $l_f=0.03$. Predicted (a) mean and (b) STD of the samples. (c) Relative error of mean and STD vs number of BNN samples. (d) Predicted covariance kernel and (e) exact covariance kernel of $\log k - 0.5$. This example demonstrates that our method can solve a high-dimensional inverse problem of SPDE.}
   \label{fig:elliptic_high}
    
\end{figure}

To apply our example to the high-dimensional inverse problem, we next solve the same equation with $l_k=0.03$ and $l_f=0.03$. To capture the oscillatory behavior of random processes, the number of sensors of each $u$ and $f$ increases, i.e., $N_u=201$ and $N_f=201$. In addition, we use Fourier embeddings with $\sigma_1=1.0$ and $\sigma_2=20.0$, and the rest of the network settings are the same as in the previous example. The number of leapfrog step is $M=\text{2,000}$ and the step size is $\delta=5 \times 10^{-6}$. The results provide good approximations (Figure \ref{fig:elliptic_high}). This example demonstrates that our method can quantify uncertainty in a high-dimensional inverse problem of SPDE.

\section{Conclusion}\label{sec:conclusion}

In this paper, we present a new data-driven method based on BNN and HMC to solve the
forward and inverse problems of SPDEs. The solutions of SPDEs are modeled by
BNN, and the distributions of the network parameters are inferred by using Bayes'
rule. The governing physical law of the SPDE is encoded into the posterior
distribution by using automatic differentiation. 
The posterior distribution is efficiently sampled using HMC to quantify
uncertainties.

We first demonstrated that our method approximate random processes well and
obtain accurate predictions of the mean, STD, and covariance. Subsequently, we
consider Poisson problems with source terms of different correlation lengths
which determine the dimension of the problems. Our framework not only provides
good predictions even for the high-dimensional problems but also has the
computational cost almost independent of the dimension of the problem showing
the potential for tackling the curse of dimensionality. In the case of 2D
Allen-Cahn equation, we show that our method works well for the nonlinear
equation. Finally, our method is applied to the inverse problem of the elliptic
equation; inferred mean and STD, and covariance of the PDE parameter agree well
with those of the reference. 

While the choice of a network structure for the BNN model may have a significant
impact on predictions, it remains an open problem to choose an optimal network
structure. In particular, we must determine hyperparameters in FFN according to
the frequency feature contained in problems, but no concrete guideline has been
established. In addition, the prior distribution may need to be chosen with care
in order to achieve more accurate prediction, which remains future work.

\section*{Acknowledgements}

This work was supported by the National Research Foundation of Korea [NRF-2021R1C1C1007875].

\bibliographystyle{unsrt}
\bibliography{ref}

\begin{thebibliography}{10}

\bibitem{zhuang2021learned}
Jiawei Zhuang, Dmitrii Kochkov, Yohai Bar-Sinai, Michael~P Brenner, and Stephan
  Hoyer.
\newblock Learned discretizations for passive scalar advection in a
  two-dimensional turbulent flow.
\newblock {\em Physical Review Fluids}, 6(6):064605, 2021.

\bibitem{bar2019learning}
Yohai Bar-Sinai, Stephan Hoyer, Jason Hickey, and Michael~P Brenner.
\newblock Learning data-driven discretizations for partial differential
  equations.
\newblock {\em Proceedings of the National Academy of Sciences},
  116(31):15344--15349, 2019.

\bibitem{chudomelka2020deep}
Bryce Chudomelka, Youngjoon Hong, Hyunwoo Kim, and Jinyoung Park.
\newblock Deep neural network for solving differential equations motivated by
  {L}egendre-{G}alerkin approximation.
\newblock {\em arXiv preprint arXiv:2010.12975}, 2020.

\bibitem{rudd2015constrained}
Keith Rudd and Silvia Ferrari.
\newblock A constrained integration ({CINT}) approach to solving partial
  differential equations using artificial neural networks.
\newblock {\em Neurocomputing}, 155:277--285, 2015.

\bibitem{raissi2020hidden}
Maziar Raissi, Alireza Yazdani, and George~Em Karniadakis.
\newblock Hidden fluid mechanics: Learning velocity and pressure fields from
  flow visualizations.
\newblock {\em Science}, 367(6481):1026--1030, 2020.

\bibitem{raissi2019physics}
Maziar Raissi, Paris Perdikaris, and George~E Karniadakis.
\newblock Physics-informed neural networks: A deep learning framework for
  solving forward and inverse problems involving nonlinear partial differential
  equations.
\newblock {\em Journal of Computational Physics}, 378:686--707, 2019.

\bibitem{kharazmi2019variational}
Ehsan Kharazmi, Zhongqiang Zhang, and George~Em Karniadakis.
\newblock Variational physics-informed neural networks for solving partial
  differential equations.
\newblock {\em arXiv preprint arXiv:1912.00873}, 2019.

\bibitem{raissi2018hidden}
Maziar Raissi and George~Em Karniadakis.
\newblock Hidden physics models: Machine learning of nonlinear partial
  differential equations.
\newblock {\em Journal of Computational Physics}, 357:125--141, 2018.

\bibitem{wang2021learning}
Sifan Wang, Hanwen Wang, and Paris Perdikaris.
\newblock Learning the solution operator of parametric partial differential
  equations with physics-informed {DeepONets}.
\newblock {\em Science advances}, 7(40):eabi8605, 2021.

\bibitem{li2020fourier}
Zongyi Li, Nikola Kovachki, Kamyar Azizzadenesheli, Burigede Liu, Kaushik
  Bhattacharya, Andrew Stuart, and Anima Anandkumar.
\newblock Fourier neural operator for parametric partial differential
  equations.
\newblock {\em arXiv preprint arXiv:2010.08895}, 2020.

\bibitem{lagaris1998artificial}
Isaac~E Lagaris, Aristidis Likas, and Dimitrios~I Fotiadis.
\newblock Artificial neural networks for solving ordinary and partial
  differential equations.
\newblock {\em IEEE transactions on neural networks}, 9(5):987--1000, 1998.

\bibitem{lagaris2000neural}
Isaac~E Lagaris, Aristidis~C Likas, and Dimitris~G Papageorgiou.
\newblock Neural-network methods for boundary value problems with irregular
  boundaries.
\newblock {\em IEEE Transactions on Neural Networks}, 11(5):1041--1049, 2000.

\bibitem{dissanayake1994neural}
MWMG Dissanayake and Nhan Phan-Thien.
\newblock Neural-network-based approximations for solving partial differential
  equations.
\newblock {\em communications in Numerical Methods in Engineering},
  10(3):195--201, 1994.

\bibitem{meade1994solution}
Andrew~J Meade~Jr and Alvaro~A Fernandez.
\newblock Solution of nonlinear ordinary differential equations by feedforward
  neural networks.
\newblock {\em Mathematical and Computer Modelling}, 20(9):19--44, 1994.

\bibitem{tensorflow2015-whitepaper}
Mart\'{\i}n Abadi, Ashish Agarwal, Paul Barham, Eugene Brevdo, Zhifeng Chen,
  Craig Citro, Greg~S. Corrado, Andy Davis, Jeffrey Dean, Matthieu Devin,
  Sanjay Ghemawat, Ian Goodfellow, Andrew Harp, Geoffrey Irving, Michael Isard,
  Yangqing Jia, Rafal Jozefowicz, Lukasz Kaiser, Manjunath Kudlur, Josh
  Levenberg, Dandelion Man\'{e}, Rajat Monga, Sherry Moore, Derek Murray, Chris
  Olah, Mike Schuster, Jonathon Shlens, Benoit Steiner, Ilya Sutskever, Kunal
  Talwar, Paul Tucker, Vincent Vanhoucke, Vijay Vasudevan, Fernanda Vi\'{e}gas,
  Oriol Vinyals, Pete Warden, Martin Wattenberg, Martin Wicke, Yuan Yu, and
  Xiaoqiang Zheng.
\newblock {TensorFlow}: Large-scale machine learning on heterogeneous systems,
  2015.
\newblock Software available from tensorflow.org.

\bibitem{NEURIPS2019_9015}
Adam Paszke, Sam Gross, Francisco Massa, Adam Lerer, James Bradbury, Gregory
  Chanan, Trevor Killeen, Zeming Lin, Natalia Gimelshein, Luca Antiga, Alban
  Desmaison, Andreas Kopf, Edward Yang, Zachary DeVito, Martin Raison, Alykhan
  Tejani, Sasank Chilamkurthy, Benoit Steiner, Lu~Fang, Junjie Bai, and Soumith
  Chintala.
\newblock Pytorch: An imperative style, high-performance deep learning library.
\newblock In H.~Wallach, H.~Larochelle, A.~Beygelzimer, F.~d\textquotesingle
  Alch\'{e}-Buc, E.~Fox, and R.~Garnett, editors, {\em Advances in Neural
  Information Processing Systems 32}, pages 8024--8035. Curran Associates,
  Inc., 2019.

\bibitem{tripathy2018deep}
Rohit~K Tripathy and Ilias Bilionis.
\newblock Deep {UQ}: Learning deep neural network surrogate models for high
  dimensional uncertainty quantification.
\newblock {\em Journal of Computational Physics}, 375:565--588, 2018.

\bibitem{zhu2019physics}
Yinhao Zhu, Nicholas Zabaras, Phaedon-Stelios Koutsourelakis, and Paris
  Perdikaris.
\newblock Physics-constrained deep learning for high-dimensional surrogate
  modeling and uncertainty quantification without labeled data.
\newblock {\em Journal of Computational Physics}, 394:56--81, 2019.

\bibitem{zhu2018bayesian}
Yinhao Zhu and Nicholas Zabaras.
\newblock Bayesian deep convolutional encoder-decoder networks for surrogate
  modeling and uncertainty quantification.
\newblock {\em Journal of Computational Physics}, 366:415--447, 2018.

\bibitem{mo2019deepa}
Shaoxing Mo, Yinhao Zhu, Nicholas Zabaras, Xiaoqing Shi, and Jichun Wu.
\newblock Deep convolutional encoder-decoder networks for uncertainty
  quantification of dynamic multiphase flow in heterogeneous media.
\newblock {\em Water Resources Research}, 55(1):703--728, 2019.

\bibitem{mo2019deepb}
Shaoxing Mo, Nicholas Zabaras, Xiaoqing Shi, and Jichun Wu.
\newblock Deep autoregressive neural networks for high-dimensional inverse
  problems in groundwater contaminant source identification.
\newblock {\em Water Resources Research}, 55(5):3856--3881, 2019.

\bibitem{yang2020physics}
Liu Yang, Dongkun Zhang, and George~Em Karniadakis.
\newblock Physics-informed generative adversarial networks for stochastic
  differential equations.
\newblock {\em SIAM Journal on Scientific Computing}, 42(1):A292--A317, 2020.

\bibitem{yang2019adversarial}
Yibo Yang and Paris Perdikaris.
\newblock Adversarial uncertainty quantification in physics-informed neural
  networks.
\newblock {\em Journal of Computational Physics}, 394:136--152, 2019.

\bibitem{guo2022normalizing}
Ling Guo, Hao Wu, and Tao Zhou.
\newblock Normalizing field flows: Solving forward and inverse stochastic
  differential equations using physics-informed flow models.
\newblock {\em Journal of Computational Physics}, page 111202, 2022.

\bibitem{zhang2019quantifying}
Dongkun Zhang, Lu~Lu, Ling Guo, and George~Em Karniadakis.
\newblock Quantifying total uncertainty in physics-informed neural networks for
  solving forward and inverse stochastic problems.
\newblock {\em Journal of Computational Physics}, 397:108850, 2019.

\bibitem{xiu2002wiener}
Dongbin Xiu and George~Em Karniadakis.
\newblock The {W}iener-{A}skey polynomial chaos for stochastic differential
  equations.
\newblock {\em SIAM Journal on Scientific Computing}, 24(2):619--644, 2002.

\bibitem{zhang2020learning}
Dongkun Zhang, Ling Guo, and George~Em Karniadakis.
\newblock Learning in modal space: Solving time-dependent stochastic {PDE}s
  using physics-informed neural networks.
\newblock {\em SIAM Journal on Scientific Computing}, 42(2):A639--A665, 2020.

\bibitem{choi2014equivalence}
Minseok Choi, Themistoklis~P Sapsis, and George~Em Karniadakis.
\newblock On the equivalence of dynamically orthogonal and bi-orthogonal
  methods: Theory and numerical simulations.
\newblock {\em Journal of Computational Physics}, 270:1--20, 2014.

\bibitem{karumuri2020simulator}
Sharmila Karumuri, Rohit Tripathy, Ilias Bilionis, and Jitesh Panchal.
\newblock Simulator-free solution of high-dimensional stochastic elliptic
  partial differential equations using deep neural networks.
\newblock {\em Journal of Computational Physics}, 404:109120, 2020.

\bibitem{nabian2018deep}
Mohammad~Amin Nabian and Hadi Meidani.
\newblock A deep neural network surrogate for high-dimensional random partial
  differential equations.
\newblock {\em arXiv preprint arXiv:1806.02957}, 2018.

\bibitem{neal2012bayesian}
Radford~M Neal.
\newblock {\em Bayesian learning for neural networks}, volume 118.
\newblock Springer Science \& Business Media, 2012.

\bibitem{graves2011practical}
Alex Graves.
\newblock Practical variational inference for neural networks.
\newblock In J.~Shawe-Taylor, R.~Zemel, P.~Bartlett, F.~Pereira, and K.Q.
  Weinberger, editors, {\em Advances in Neural Information Processing Systems},
  volume~24. Curran Associates, Inc., 2011.

\bibitem{blundell2015weight}
Charles Blundell, Julien Cornebise, Koray Kavukcuoglu, and Daan Wierstra.
\newblock Weight uncertainty in neural network.
\newblock In {\em Proceedings of the 32nd International Conference on Machine
  Learning}, pages 1613--1622. PMLR, 2015.

\bibitem{gal2016dropout}
Yarin Gal and Zoubin Ghahramani.
\newblock Dropout as a bayesian approximation: Representing model uncertainty
  in deep learning.
\newblock In {\em Proceedings of The 33rd International Conference on Machine
  Learning}, pages 1050--1059. PMLR, 2016.

\bibitem{welling2011bayesian}
Max Welling and Yee~Whye Teh.
\newblock Bayesian learning via stochastic gradient {L}angevin dynamics.
\newblock In {\em Proceedings of The 28th International Conference on Machine
  Learning}, pages 681--688, 2011.

\bibitem{liu2016stein}
Qiang Liu and Dilin Wang.
\newblock Stein variational gradient descent: A general purpose bayesian
  inference algorithm.
\newblock {\em arXiv preprint arXiv:1608.04471}, 2016.

\bibitem{yang2021b}
Liu Yang, Xuhui Meng, and George~Em Karniadakis.
\newblock {B-PINNs}: Bayesian physics-informed neural networks for forward and
  inverse {PDE} problems with noisy data.
\newblock {\em Journal of Computational Physics}, 425:109913, 2021.

\bibitem{sun2020physics}
Luning Sun and Jian-Xun Wang.
\newblock Physics-constrained bayesian neural network for fluid flow
  reconstruction with sparse and noisy data.
\newblock {\em Theoretical and Applied Mechanics Letters}, 10(3):161--169,
  2020.

\bibitem{duane1987hybrid}
Simon Duane, Anthony~D Kennedy, Brian~J Pendleton, and Duncan Roweth.
\newblock Hybrid {M}onte {C}arlo.
\newblock {\em Physics letters B}, 195(2):216--222, 1987.

\bibitem{betancourt2017conceptual}
Michael Betancourt.
\newblock A conceptual introduction to {H}amiltonian {M}onte {C}arlo.
\newblock {\em arXiv preprint arXiv:1701.02434}, 2017.

\bibitem{parzen1962estimation}
Emanuel Parzen.
\newblock On estimation of a probability density function and mode.
\newblock {\em The annals of mathematical statistics}, 33(3):1065--1076, 1962.

\bibitem{silverman2018density}
Bernard~W Silverman.
\newblock {\em Density estimation for statistics and data analysis}.
\newblock Routledge, 2018.

\bibitem{rezende2015variational}
Danilo Rezende and Shakir Mohamed.
\newblock Variational inference with normalizing flows.
\newblock In {\em Proceedings of the 32nd International Conference on Machine
  Learning}, pages 1530--1538. PMLR, 2015.

\bibitem{dinh2016density}
Laurent Dinh, Jascha Sohl-Dickstein, and Samy Bengio.
\newblock Density estimation using {R}eal {NVP}.
\newblock {\em arXiv preprint arXiv:1605.08803}, 2016.

\bibitem{papamakarios2017masked}
George Papamakarios, Theo Pavlakou, and Iain Murray.
\newblock Masked autoregressive flow for density estimation.
\newblock In I.~Guyon, U.~Von Luxburg, S.~Bengio, H.~Wallach, R.~Fergus,
  S.~Vishwanathan, and R.~Garnett, editors, {\em Advances in Neural Information
  Processing Systems}, volume~30. Curran Associates, Inc., 2017.

\bibitem{kingma2016improved}
Durk~P Kingma, Tim Salimans, Rafal Jozefowicz, Xi~Chen, Ilya Sutskever, and Max
  Welling.
\newblock Improved variational inference with inverse autoregressive flow.
\newblock {\em Advances in neural information processing systems},
  29:4743--4751, 2016.

\bibitem{paalanen2006feature}
Pekka Paalanen, Joni-Kristian Kamarainen, Jarmo Ilonen, and Heikki
  K{\"a}lvi{\"a}inen.
\newblock {F}eature representation and discrimination based on {G}aussian
  mixture model probability densities—{P}ractices and algorithms.
\newblock {\em Pattern Recognition}, 39(7):1346--1358, 2006.

\bibitem{bilmes1998gentle}
Jeff Bilmes.
\newblock A gentle tutorial of the {EM} algorithm and its application to
  parameter estimation for {G}aussian mixture and hidden {M}arkov models.
\newblock Technical Report TR-97-021, International Computer Science Institute,
  1997.

\bibitem{yu2011solving}
Guoshen Yu, Guillermo Sapiro, and St{\'e}phane Mallat.
\newblock Solving inverse problems with piecewise linear estimators: From
  {G}aussian mixture models to structured sparsity.
\newblock {\em IEEE Transactions on Image Processing}, 21(5):2481--2499, 2011.

\bibitem{beskos2013optimal}
Alexandros Beskos, Natesh Pillai, Gareth Roberts, Jesus-Maria Sanz-Serna, and
  Andrew Stuart.
\newblock Optimal tuning of the hybrid {M}onte {C}arlo algorithm.
\newblock {\em Bernoulli}, 19(5A):1501--1534, 2013.

\bibitem{mangoubi2018dimensionally}
Oren Mangoubi and Nisheeth Vishnoi.
\newblock Dimensionally tight bounds for second-order {H}amiltonian {M}onte
  {C}arlo.
\newblock In S.~Bengio, H.~Wallach, H.~Larochelle, K.~Grauman, N.~Cesa-Bianchi,
  and R.~Garnett, editors, {\em Advances in Neural Information Processing
  Systems}, volume~31. Curran Associates, Inc., 2018.

\bibitem{tancik2020fourier}
Matthew Tancik, Pratul Srinivasan, Ben Mildenhall, Sara Fridovich-Keil, Nithin
  Raghavan, Utkarsh Singhal, Ravi Ramamoorthi, Jonathan Barron, and Ren Ng.
\newblock Fourier features let networks learn high frequency functions in low
  dimensional domains.
\newblock {\em Advances in Neural Information Processing Systems},
  33:7537--7547, 2020.

\bibitem{rahaman2019spectral}
Nasim Rahaman, Aristide Baratin, Devansh Arpit, Felix Draxler, Min Lin, Fred
  Hamprecht, Yoshua Bengio, and Aaron Courville.
\newblock On the spectral bias of neural networks.
\newblock In {\em Proceedings of the 36th International Conference on Machine
  Learning}, pages 5301--5310. PMLR, 2019.

\bibitem{cao2019towards}
Yuan Cao, Zhiying Fang, Yue Wu, Ding-Xuan Zhou, and Quanquan Gu.
\newblock Towards understanding the spectral bias of deep learning.
\newblock {\em arXiv preprint arXiv:1912.01198}, 2019.

\bibitem{ronen2019convergence}
Basri Ronen, David Jacobs, Yoni Kasten, and Shira Kritchman.
\newblock The convergence rate of neural networks for learned functions of
  different frequencies.
\newblock In H.~Wallach, H.~Larochelle, A.~Beygelzimer, F.~d\textquotesingle
  Alch\'{e}-Buc, E.~Fox, and R.~Garnett, editors, {\em Advances in Neural
  Information Processing Systems}, volume~32. Curran Associates, Inc., 2019.

\bibitem{wang2021eigenvector}
Sifan Wang, Hanwen Wang, and Paris Perdikaris.
\newblock On the eigenvector bias of {F}ourier feature networks: From
  regression to solving multi-scale {PDE}s with physics-informed neural
  networks.
\newblock {\em Computer Methods in Applied Mechanics and Engineering},
  384:113938, 2021.

\bibitem{jacot2018neural}
Arthur Jacot, Franck Gabriel, and Clement Hongler.
\newblock Neural tangent kernel: Convergence and generalization in neural
  networks.
\newblock In S.~Bengio, H.~Wallach, H.~Larochelle, K.~Grauman, N.~Cesa-Bianchi,
  and R.~Garnett, editors, {\em Advances in Neural Information Processing
  Systems}, volume~31. Curran Associates, Inc., 2018.

\bibitem{arora2019exact}
Sanjeev Arora, Simon~S Du, Wei Hu, Zhiyuan Li, Russ~R Salakhutdinov, and
  Ruosong Wang.
\newblock On exact computation with an infinitely wide neural net.
\newblock In H.~Wallach, H.~Larochelle, A.~Beygelzimer, F.~d\textquotesingle
  Alch\'{e}-Buc, E.~Fox, and R.~Garnett, editors, {\em Advances in Neural
  Information Processing Systems}, volume~32. Curran Associates, Inc., 2019.

\bibitem{lee2019wide}
Jaehoon Lee, Lechao Xiao, Samuel Schoenholz, Yasaman Bahri, Roman Novak, Jascha
  Sohl-Dickstein, and Jeffrey Pennington.
\newblock Wide neural networks of any depth evolve as linear models under
  gradient descent.
\newblock In H.~Wallach, H.~Larochelle, A.~Beygelzimer, F.~d\textquotesingle
  Alch\'{e}-Buc, E.~Fox, and R.~Garnett, editors, {\em Advances in Neural
  Information Processing Systems}, volume~32. Curran Associates, Inc., 2019.

\bibitem{logg2012automated}
Anders Logg, Kent-Andre Mardal, and Garth Wells.
\newblock {\em Automated solution of differential equations by the finite
  element method: The {FEniCS} book}, volume~84.
\newblock Springer Science \& Business Media, 2012.

\bibitem{alnaes2015fenics}
Martin Aln{\ae}s, Jan Blechta, Johan Hake, August Johansson, Benjamin Kehlet,
  Anders Logg, Chris Richardson, Johannes Ring, Marie~E Rognes, and Garth~N
  Wells.
\newblock The {FEniCS} project version 1.5.
\newblock {\em Archive of Numerical Software}, 3(100), 2015.

\end{thebibliography}

\end{document}